%% file: main.tex
\pdfoutput=1

\documentclass{article}

\usepackage[preprint,nonatbib]{neurips_2025}

\usepackage{tabularx}

\input{preamble}

\usepackage[colorlinks,citecolor=blue,linkcolor=black,urlcolor=blue]{hyperref}

\title{A Survey on Adversarial Attacks and Defenses for Diffusion Models Across Multiple Modalities}

\author{%
  \bfseries
  Ozgur Kara\textsuperscript{1}\quad
  Tarik Can Ozden\textsuperscript{1}\quad
  Furkan Horoz\textsuperscript{1}\quad
  Zeqian Long\textsuperscript{1,3}\quad
  Haotian Xue\textsuperscript{2}\\
  \bfseries
  Yipu Chen\textsuperscript{2}\quad
  Oguzhan Akcin\textsuperscript{4}\quad
  Yongxin Chen\textsuperscript{2}\quad
  James Matthew Rehg\textsuperscript{1}\\[0.7em]
  \normalfont\textsuperscript{1}University of Illinois Urbana-Champaign\quad
  \textsuperscript{2}Georgia Institute of Technology\\
  \normalfont\textsuperscript{3}Stanford University\quad
  \textsuperscript{4}The University of Texas at Austin%
}

\begin{document}

\maketitle

\begin{abstract}
Diffusion models have become the dominant family of generative models in the visual domain. However, their widespread public availability enables misuse at scale, motivating a rapidly growing body of research on adversarial attacks and defenses. This survey provides, to our knowledge, the first unified review of this literature across three visual modalities: image, video, and 3D. We introduce a comprehensive, task-centric taxonomy: we first divide the literature by modality; within each modality, we separate methods into attacks and defenses, and then group them by the generative task they target, presenting them chronologically within each task. Moreover, we provide an in-depth analysis of their evaluation settings, consolidating the datasets, metrics, and benchmarks used to assess them. We conclude by identifying several open challenges and outlining concrete future research directions. \textbf{Project Webpage}: \url{https://github.com/ozgurkara99/awesome-adv-attack-defense-on-diffusion}.
\end{abstract}

\begin{center}
  \textbf{Keywords:} Diffusion models \(\cdot\) Adversarial attacks \(\cdot\) Adversarial defenses
\end{center}

\input{sections/01_introduction}

\input{sections/02_preliminaries}

\input{sections/03_attacks_defenses}

\input{sections/04_evaluation}

\input{sections/05_open_challenges_conclusion}

\bibliographystyle{splncs04}
\bibliography{main}

\end{document}

%% file: preamble.tex
\usepackage[T1]{fontenc}
\usepackage[dvipsnames]{xcolor}
\usepackage{amsmath}
\usepackage{amssymb}
\usepackage{cite}          
\usepackage[hyphens]{url}
\usepackage[labelfont=bf,font=small,tableposition=bottom]{caption}
\usepackage[skip=3pt]{subcaption}
\usepackage{graphicx}
\usepackage{booktabs}
\usepackage{array}         
\usepackage{afterpage}     
\usepackage{longtable}     
\usepackage{colortbl}      
\usepackage{pifont}        
\usepackage{adjustbox}     
\usepackage{multirow}
\usepackage{tikz}          
\usetikzlibrary{arrows.meta,positioning,shadows}
\usepackage[edges]{forest} 
\usepackage{eccvabbrv}
\usepackage{xspace}
\newcommand{\dms}{DMs\xspace}
\newcommand{\ddpms}{DDPMs\xspace}

\newcommand{\dit}{DiT\xspace}      
\newcommand{\dits}{DiTs\xspace}
\newcommand{\vtov}{V2V\xspace}     
\newcommand{\gsplat}{3DGS\xspace}  

\newcommand{\cmark}{\textcolor{green!60!black}{\ding{51}}}
\newcommand{\xmark}{\textcolor{red!70!black}{\ding{55}}}
\definecolor{atkcol}{RGB}{175,45,45}  
\definecolor{defcol}{RGB}{30,110,75}  
\colorlet{bandDef}{defcol!22}         
\definecolor{sideAtk}{RGB}{223,240,216}  
\definecolor{sideCoop}{RGB}{225,225,225} 
\definecolor{sideDef}{RGB}{252,229,205}  
\definecolor{bandImg}{RGB}{197,217,241}  
\definecolor{bandVid}{RGB}{244,204,204}  
\definecolor{band3D}{RGB}{214,203,231}   
\definecolor{oursrow}{RGB}{255,242,204}  
\colorlet{rowshade}{gray!13}             
\newcommand{\rowstrut}{\rule[-0.9ex]{0pt}{3.2ex}}
\newlength{\tblrest}
\newcommand{\grouprow}[2]{\multicolumn{4}{@{}l@{}}{\cellcolor{#1}\textbf{#2}}}

\newcommand{\grouprowImg}[2]{%
  \midrule
  \rowcolor{#1}%
  \multicolumn{8}{@{}l@{}}{\rowstrut\textbf{#2}}\\
}
\definecolor{ozgurcolor}{RGB}{230,97,0}   

\makeatletter
\renewcommand\subsubsection{\@startsection{subsubsection}{3}{\z@}%
  {-7\p@ \@plus -2\p@ \@minus -2\p@}%
  {4\p@}%
  {\normalfont\normalsize\bfseries\itshape}}
\makeatother

%% file: sections/01_introduction.tex
\section{Introduction}
\label{sec:introduction}
Diffusion models (\dms)~\cite{ho2020denoising,sohl2015deep,rombach2022high} have surpassed GANs~\cite{goodfellow2020generative, radford2015unsupervised, arjovsky2017wasserstein, karras2019style, mirza2014conditional} to become the dominant family of deep generative models, expanding from high-fidelity image synthesis~\cite{dhariwal2021diffusion,rombach2022high} to video~\cite{singermake,ho2022imagen,yang_cogvideox_2024,peng_controlnext_2025, ma2025controllable} and 3D content generation~\cite{pooledreamfusion,xiang2025repurposing,yi2024gaussiandreamer,zhuang2023dreameditor,wang2024gaussianeditor}, while their architectures evolved from U-Nets~\cite{ronneberger2015u} to diffusion transformers (\dits)~\cite{peebles2023scalable}. Now publicly available as open-source releases and commercial services~\cite{rombach2022high,midjourney2026,openai2023dalle3,openai2024sora,kuaishou2024kling}, \dms place unprecedented generative power in anyone's hands, making misuse a threat at scale. Because each modality supports its own generative tasks, every task opens a distinct attack surface: images face unauthorized personalization, style transfer, and face swapping; videos, manipulated talking-head animations and deepfake editing; and 3D assets, instruction-driven scene alterations. These harms were first met \emph{reactively}, through deepfake detection and watermark-based authentication, but such post-hoc measures are easily circumvented, and the field has since turned \emph{proactive}, aiming to make content inherently resistant to exploitation before it is misused. This shift defines the literature we survey: \emph{adversarial attacks and defenses for diffusion models}.

\input{figures/fig1_survey_organization}

Adversarial attacks and defenses for \dms differ fundamentally from their counterparts on classification models, introducing novel challenges. \emph{(1)}~\textbf{Task and setting heterogeneity}: Whereas a classifier presents a single, well-defined decision boundary, generative pipelines vary widely in task, input modality, and backbone architecture, so protections crafted for one setting transfer poorly to another. \emph{(2)}~\textbf{Multi-step, conditional, and stochastic generation}: Instead of fooling one deterministic forward pass, a perturbation must remain effective across an iterative denoising trajectory with random noise draws and user-chosen conditioning, and optimizing through this long chain is far more expensive than a one-step attack. \emph{(3)}~\textbf{An adaptive adversary}: Uniquely, the party deploying the model often acts as the adversary; it controls the entire pipeline and can preprocess inputs or modify the model itself to strip or circumvent protections, so protections must hold against a moving target rather than a fixed one.

This literature has grown quickly but remains fragmented, and no existing survey covers all visual modalities. Table~\ref{tab:survey_comparison} positions our work against nine related surveys. Notably, none address 3D content, making ours the first to jointly cover image, video, and 3D with a primary focus on adversarial robustness. Existing general adversarial-robustness surveys for \dms remain exclusively image-centric~\cite{zhang2025surveyimmunization,peng2024protective,truong2025attacks,zhang2025adversarial,wei2025responsiblediffusion,alotaibi2026adversarial}.
The surveys that do reach video approach it strictly through facial deepfakes and the detection or disruption of AI-generated media~\cite{nguyenle2025proactive,deng2025aigcdefenses}. The remaining survey studies replication, trustworthiness, copyright, and watermarking, where adversarial robustness is at most peripheral~\cite{wang2024replication}.

\afterpage{\input{tables/table1_survey_comparison}}

Beyond coverage, the deeper problem is organization. Prior taxonomies sort methods by perturbation space or threat category---axes that describe \emph{how} a method operates, but not \emph{what} it protects against---leaving works that target the same misuse scattered and incomparable. In practice, both misuse and protection are task-specific: a diffusion model is attacked through a concrete generative task, and each modality imposes its own structural constraints---precisely why methods transfer poorly across tasks, modalities, and backbones. We therefore organize the literature first by modality, then into attacks and defenses, and finally by the generative task they target, ordered chronologically within each task. This makes methods addressing the same threat directly comparable, exposes each task's attack/counterattack arms race, and localizes modality-specific challenges where they belong. Figure~\ref{fig:structure} summarizes the resulting organization.

To summarize, we provide (i)~the first unified review of adversarial attacks and defenses for \dms covering all visual domains, namely image, video and 3D, under a single immunization framework (Section~\ref{subsec:attacks_threat_model}); (ii)~a \emph{task-centric} taxonomy that categorizes methods first by modality and then by the generative task they attack or protect (Figure~\ref{fig:hierarchy}), making methods that target the same task directly comparable; (iii)~a consolidated treatment of the datasets, metrics, dedicated benchmarks, and analyses used to evaluate them (Section~\ref{sec:evaluation}; Tables~\ref{tab:methods_image}, \ref{tab:methods_video}, \ref{tab:methods_3d}); and (iv)~an outline of open challenges with concrete future research directions (Section~\ref{sec:open_challenges}).

%% file: figures/fig1_survey_organization.tex
\definecolor{structcol}{RGB}{55,75,105}
\definecolor{prelimBg}{RGB}{20,110,115}   
\definecolor{prelimFw}{RGB}{170,110,15}   
\begin{figure}[tbp]
\centering
\adjustbox{max width=\linewidth}{%
\begin{forest}
  forked edges,
  for tree={
    font=\sffamily,
    grow'=east,
    parent anchor=east,
    child anchor=west,
    anchor=west,
    calign=center,
    edge={line width=0.8pt, structcol!70},
    rounded corners=3pt,
    inner xsep=6pt,
    inner ysep=4pt,
    l sep=5mm,
    s sep=1.5mm,
    fork sep=3mm
  },
  where n children=0{tier=desc}{},
  S/.style={font=\sffamily\bfseries\large,align=center,rounded corners=3pt,fill=structcol!15,draw=structcol!50,line width=1pt},
  L/.style={font=\sffamily\normalsize,rounded corners=2pt,fill=structcol!4,draw=structcol!25},
  Ub/.style={font=\sffamily\bfseries\normalsize,align=center,fill=prelimBg!15,draw=prelimBg!40},
  Lb/.style={font=\sffamily\normalsize,rounded corners=2pt,fill=prelimBg!5,draw=prelimBg!25},
  Uf/.style={font=\sffamily\bfseries\normalsize,align=center,fill=prelimFw!15,draw=prelimFw!40},
  Lf/.style={font=\sffamily\normalsize,rounded corners=2pt,fill=prelimFw!5,draw=prelimFw!25},
  Ui/.style={font=\sffamily\bfseries\normalsize,align=center,fill=blue!15,draw=blue!40},
  Uv/.style={font=\sffamily\bfseries\normalsize,align=center,fill=red!15,draw=red!40},
  Ug/.style={font=\sffamily\bfseries\normalsize,align=center,fill=violet!15,draw=violet!40},
  Li/.style={font=\sffamily\normalsize,rounded corners=2pt,fill=blue!5,draw=blue!25},
  Lv/.style={font=\sffamily\normalsize,rounded corners=2pt,fill=red!5,draw=red!25},
  Lg/.style={font=\sffamily\normalsize,rounded corners=2pt,fill=violet!5,draw=violet!25},
[, phantom, for children={no edge}
  [{\S\ref{sec:introduction}~Introduction}, S
    [{\parbox{10.2cm}{\raggedright Motivation for proactive content immunization, unique adversarial challenges of generative pipelines, and survey contributions.}}, L]]
  [{\S\ref{sec:preliminaries}~Preliminaries}, S
    [{\S\ref{subsec:frameworks}~Background on \dms}, Ub
      [{\parbox{10.2cm}{\raggedright Core mechanics of diffusion modeling, architectural vulnerabilities, and structural constraints per vision modality.}}, Lb]]
    [{\S\ref{subsec:attacks_threat_model}~A Unified Framework}, Uf
      [{\parbox{10.2cm}{\raggedright A generalized formulation of the threat model, encompassing proactive cloaking and adaptive countermeasures.}}, Lf]]
  ]
  [{\S\ref{sec:attacks_defenses}~Adversarial Attacks \& Defenses}, S
    [{\S\ref{subsec:image}~Image}, Ui
      [{\parbox{10.2cm}{\raggedright Categorization of attacks targeting various 2D image tasks and defensive strategies like purification and model unlearning. \hfill (Table~\ref{tab:methods_image})}}, Li]]
    [{\S\ref{subsec:video}~Video}, Uv
      [{\parbox{10.2cm}{\raggedright Analysis of vulnerabilities in temporal generation and motion-editing tasks alongside nascent video-specific defenses. \hfill (Table~\ref{tab:methods_video})}}, Lv]]
    [{\S\ref{subsec:3d}~3D}, Ug
      [{\parbox{10.2cm}{\raggedright Adversarial challenges in multi-view consistency, spatial editing, and computational poisoning for 3D representations. \hfill (Table~\ref{tab:methods_3d})}}, Lg]]
  ]
  [{\S\ref{sec:evaluation}~Evaluation}, S
    [{\parbox{10.2cm}{\raggedright Consolidated datasets and benchmarks, target models, and evaluation metrics across the surveyed literature. \hfill (Tables~\ref{tab:dataset_overview}, \ref{tab:metric_definitions})}}, L]]
  [{\S\ref{sec:open_challenges}~Open Challenges \& Conclusion}, S
    [{\parbox{10.2cm}{\raggedright Key research directions: breaking the purification arms race, achieving robust generalization, and expanding theoretical foundations.}}, L]]
]
\end{forest}}%
\caption{\textbf{Organization of this survey.} After motivating the problem (\S\ref{sec:introduction}) and establishing background and a unified adversarial framework (\S\ref{sec:preliminaries}), we review attacks and defenses by their target modality (\S\ref{sec:attacks_defenses}), consolidate their evaluation settings (\S\ref{sec:evaluation}), and conclude by outlining high-leverage directions for future research (\S\ref{sec:open_challenges}).}
\label{fig:structure}
\end{figure}
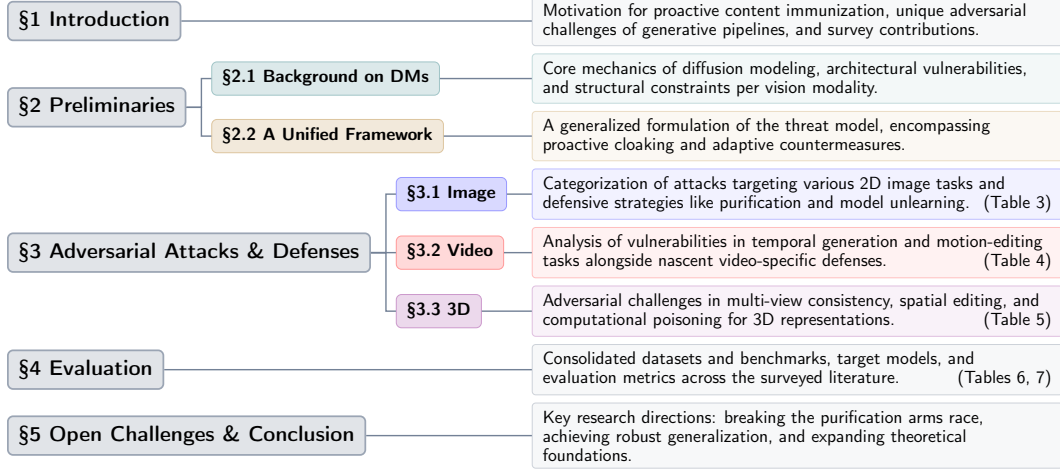

%% file: tables/table1_survey_comparison.tex
\begin{table}[!t]
\centering
{\fontsize{6.5pt}{7.8pt}\selectfont
\setlength{\tabcolsep}{2.0pt}
\renewcommand{\arraystretch}{1.15}
\caption{\textbf{Comparison with existing surveys.} \cmark\ marks a modality that is a primary subject, \xmark\ peripheral or absent; venue years give the first public release. Ours is the only survey covering image, video, and 3D together.}
\label{tab:survey_comparison}
\begin{tabularx}{\linewidth}{@{} >{\raggedright\arraybackslash}p{2.70cm} >{\centering\arraybackslash}p{2.10cm} >{\raggedright\arraybackslash}X >{\centering\arraybackslash}p{0.85cm} >{\centering\arraybackslash}p{0.85cm} >{\centering\arraybackslash}p{0.85cm} @{}}
\toprule
\multirow{2}{*}[-2pt]{\textbf{Survey}} &
\multirow{2}{*}[-2pt]{\centering\textbf{Venue}} &
\multirow{2}{*}[-2pt]{\textbf{Taxonomy Structure}} &
\multicolumn{3}{c}{\textbf{Modality Coverage}} \\
\cmidrule(lr){4-6}
 & & & \textbf{Image} & \textbf{Video} & \textbf{3D} \\
\midrule
Wang et al.~\cite{wang2024replication}
& arXiv'24 & unveiling/understanding/mitigating
& \cmark & \xmark & \xmark \\
Peng et al.~\cite{peng2024protective}
& IEEE CBD'24 & optimization objective/task
& \cmark & \xmark & \xmark \\
Truong et al.~\cite{truong2025attacks}
& ACM CSUR'25 & attack/defense
& \cmark & \xmark & \xmark \\
C.~Zhang et al.~\cite{zhang2025adversarial}
& Inf.\ Fusion'25 & target/knowledge/perturbation
& \cmark & \xmark & \xmark \\
Wei et al.~\cite{wei2025responsiblediffusion}
& arXiv'25 & threats in diffusion models
& \cmark & \xmark & \xmark \\
H.~Zhang et al.~\cite{zhang2025surveyimmunization}
& IEEE CSCloud'25 & GAN/diffusion based methods
& \cmark & \xmark & \xmark \\
Nguyen-Le et al.~\cite{nguyenle2025proactive}
& ACM CSUR'25 & disruption/watermarking
& \cmark & \cmark & \xmark \\
Deng et al.~\cite{deng2025aigcdefenses}
& ACM CSUR'25 & detection/disruption/authentication
& \cmark & \cmark & \xmark \\
Alotaibi et al.~\cite{alotaibi2026adversarial}
& arXiv'26 & diffusion model roles
& \cmark & \xmark & \xmark \\
\midrule
\rowcolor{oursrow}
\textbf{Ours}
& N/A
& \textbf{modality/generative task}
& \textbf{\cmark} & \textbf{\cmark} & \textbf{\cmark} \\
\bottomrule
\end{tabularx}}
\end{table}

%% file: sections/02_preliminaries.tex
\section{Preliminaries}
\label{sec:preliminaries}

This section provides the necessary background: Section~\ref{subsec:frameworks} reviews diffusion modeling, dominant architectures, and modality-specific attack surfaces. Section~\ref{subsec:attacks_threat_model} then formalizes a unified adversarial framework spanning both attacks and defenses.

\subsection{Background on Diffusion Models}
\label{subsec:frameworks}

\paragraph{Diffusion modeling.}
Diffusion models generate data by learning to invert a gradual, stochastic corruption process. In denoising diffusion probabilistic models (\ddpms)~\cite{ho2020denoising,sohl2015deep}, a forward process progressively adds Gaussian noise to a clean sample $\mathbf{x}_0$, yielding $\mathbf{x}_t=\sqrt{\bar{\alpha}_t}\mathbf{x}_0+\sqrt{1-\bar{\alpha}_t}\boldsymbol{\epsilon}$. A neural network $\boldsymbol{\epsilon}_\theta$ is trained to reverse this process by predicting the injected noise at each timestep $t$:

\begin{equation}
    L_{\mathrm{simple}}=\mathbb{E}_{t,\mathbf{x}_0,\boldsymbol{\epsilon}}\!\left[\|\boldsymbol{\epsilon}-\boldsymbol{\epsilon}_\theta(\sqrt{\bar{\alpha}_t}\mathbf{x}_0+\sqrt{1-\bar{\alpha}_t}\boldsymbol{\epsilon},\,t)\|^2\right].
\label{eq:lsimple}
\end{equation}

Generation iteratively applies this learned denoiser to pure noise. By conditioning the reverse process on auxiliary inputs, \dms offer a versatile generative framework. Equation~\eqref{eq:lsimple} is the core learning objective and identically the target most adversarial cloaks seek to maximize: a sample the model cannot accurately denoise is one it can neither learn from nor faithfully manipulate, effectively paralyzing the generative pipeline.

\paragraph{Model architectures.}
The denoiser's architecture dictates how information is processed and subsequently attacked. \emph{U-Net-based} \cite{blattmann2023stable, rombach2022high} models rely on a convolutional encoder-decoder, injecting conditioning signals through localized \emph{cross-attention} layers. Adversaries typically exploit this by targeting specific convolutional features or disrupting cross-attention maps. Conversely, \emph{\dit-based} models~\cite{peebles2023scalable}---powering modern state-of-the-art pipelines~\cite{yang_cogvideox_2024,openai2024sora,wan2025wan, hacohen2024ltx, labs2025flux}---replace the U-Net with a transformer backbone. DiTs operate on flattened patches, merging spatial, temporal, and conditioning tokens into a single self-attention sequence. This shift invalidates many U-Net cloaks, as global self-attention scatters a perturbation's localized impact, requiring attackers to design architecture-aware perturbations tailored specifically to DiT token-mixing.

\paragraph{Modalities.}
\dms permeate the three vision domains surveyed in this work, each introducing distinct vulnerabilities. \emph{Images} are the native setting; a single backbone supports numerous tasks, making it the most heavily studied attack surface. \emph{Video} \dms extend image backbones with temporal attention or spatio-temporal patches~\cite{blattmann2023stable,xu_hallo_2024,yang_cogvideox_2024}. This complicates protection: per-frame cloaks dilute over time, cross-frame attention mixes features, and aggressive compression codecs strip non-robust perturbations. Finally, \emph{3D} representations~\cite{mildenhall2021nerf,wang2018pixel2mesh,kerbl20233d} are not \dms themselves, but modern 3D pipelines rely on 2D \dms as priors or guidance. Adversarial methods here must account for multi-view consistency, defining budgets over rendered views or directly within the 3D parameters.

\input{figures/fig2_immunization_pipeline}

\subsection{A Unified Adversarial Framework}
\label{subsec:attacks_threat_model}

The field involves three recurring actors (Figure~\ref{fig:modality}). \textbf{Alice} (content owner) publishes content $x$ within a modality-specific domain $\mathcal{X}$. \textbf{Eve} (adversary) runs a diffusion pipeline $D_\theta$, with parameters $\theta$ and conditioning signal $c$, on Alice's content to produce unauthorized derivatives. \textbf{Bob} (defender) protects Alice by applying an imperceptible perturbation $\delta$ (the \emph{cloak}) to the content before release. We formalize these interactions below.

\paragraph{Eve's pipeline: a sampling view.}
Let $x\in\mathcal{X}$. Eve's iterative denoiser operates in a generalized form, $y = D_\theta(z, x, c)$ with $z\sim\mathcal{N}(\mathbf{0},\mathbf{I})$. This pipeline composes one-step denoisers $f_\theta^{(t)}$ from $t{=}T$ to $0$, utilizing $x$ as dictated by the generative task: e.g., a \emph{starting latent} for noise-and-denoise editing, a \emph{conditioning signal} for condition-driven generation, or \emph{training data} for model personalization. For fine-tuning tasks, Eve's pipeline adapts parameters $\Theta^\star$ (encapsulating updated model weights and/or learned embeddings) by minimizing a generic generative learning loss:

\begin{equation}
    \Theta^\star \in \arg\min_{\Theta} \mathcal{L}_{\mathrm{diff}}(\Theta; \phi(x), c).
\label{eq:ft}
\end{equation}

Here, $\mathcal{L}_{\mathrm{diff}}$ abstracts the core training objective, and $\phi:\mathcal{X}\!\to\!\mathcal{Z}$ is a modality-specific feature map carrying Alice's content into the denoiser's operating space (e.g., an identity mapping for pixels, a frozen encoder for latent spaces, or a differentiable renderer for 3D representations).

\input{tables/table2_threat_model}

\paragraph{The attack: content-level immunization.}
An attack is Bob's proactive cloak: a perturbation $\delta$ applied to Alice's content, bounded by a budget $\|\delta\|_{\mathcal{X}}\!\le\!\epsilon$, where $\|\cdot\|_{\mathcal{X}}$ is a modality-specific norm on $\mathcal{X}$ (Table~\ref{tab:threat_model_norms}). While typically an additive noise mask on 2D pixels or frames, it can take other structural forms, such as spatial or opacity offsets in 3D Gaussian Splatting. To capture this versatility, we define a generalized injection operator $\oplus$, where $x \oplus \delta$ denotes the cloaked content. The objective is to ensure Eve's pipeline produces a degraded or misdirected result rather than a faithful derivative. Formally, $\delta$ maximizes a task loss against the clean output $y^{\mathrm{clean}}{=}D_\theta(z,x,c)$:
\begin{equation}
    \delta^\star \;\in\;\arg\max_{\|\delta\|_{\mathcal{X}}\le\epsilon}\;
    \mathcal{L}_{\mathrm{task}}\!\left(D_\theta(z,\,x\oplus\delta,\,c),\, y^{\mathrm{clean}}\right).
\label{eq:bob_cloak}
\end{equation}
The design of $\mathcal{L}_{\mathrm{task}}$ adapts to the misuse: maximizing pixel-level deviation for malicious editing, derailing fine-tuned parameters $\Theta^\star(\delta)$ in~\eqref{eq:ft} for personalization, or collapsing feature representations for face-swapping. By pushing features out of the expected distribution, the cloak neutralizes the targeted generative capability.

\paragraph{The defense: adaptive countermeasures and system protections.}
Symmetrically, a defense encompasses reactive countermeasures against cloaks and broader system-level protections. First, Eve (acting as an adaptive adversary) may apply a \emph{purification} transformation, $x\oplus\delta\mapsto P(x\oplus\delta)$, using lossy compression, blurring, or diffusion-based regeneration to strip the perturbation. To survive, Bob's cloak must hold against an entire family of purifiers $\mathcal{P}$, forming a robust minimax optimization:
\begin{equation}
    \delta^\star \;\in\;\arg\max_{\|\delta\|_{\mathcal{X}}\le\epsilon}\;\min_{P\in\mathcal{P}}\;
    \mathcal{L}_{\mathrm{task}}\!\left(D_\theta(z,\, P(x\oplus\delta),\, c),\, y^{\mathrm{clean}}\right).
\label{eq:minimax}
\end{equation}
Second, platform providers can deploy system-side defenses to inherently prevent misuse: \emph{model-side} unlearning ($\theta\!\mapsto\!\tilde\theta$) to permanently erase unsafe concepts or decouple biometric identities, and \emph{conditioning-side} moderation ($c\!\mapsto\!\tilde c$) to sanitize malicious prompts prior to generation. 

\paragraph{The same framework, different modalities.}
While~\eqref{eq:bob_cloak} and~\eqref{eq:minimax} are modality-agnostic, their instantiations are not. The cloak's space, budget $\epsilon$, and entry point $\phi$ change fundamentally with the domain (Table~\ref{tab:threat_model_norms}). Because the denoiser only sees $\phi(x\oplus\delta)$, a perturbation potent at one entry point is frequently absorbed at another. This bottleneck explains why cloaks transfer poorly across backbones, necessitating modality-specific adversarial designs.

%% file: figures/fig2_immunization_pipeline.tex
\begin{figure}[t]
    \centering
    \includegraphics[width=\linewidth]{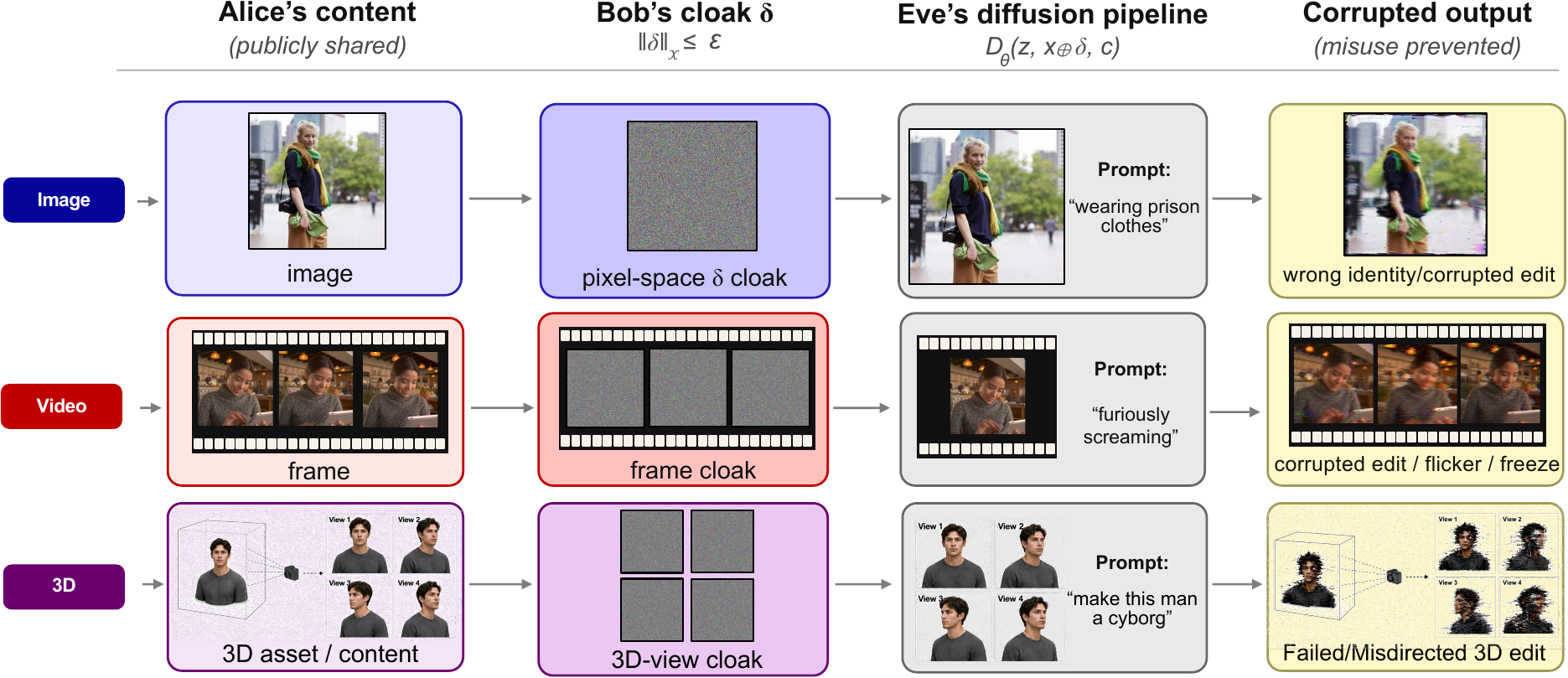}
    \caption{\textbf{Diffusion-model immunization across modalities.} Each row represents an Alice--Eve--Bob scenario: Alice publishes content $x$ (col.~1); Bob applies an imperceptible cloak $\delta$ under budget $\|\delta\|_{\mathcal{X}}\!\le\!\epsilon$ (col.~2); Eve executes her generative pipeline $D_\theta(z,\,x\oplus\delta,\,c)$ on the protected asset (col.~3), yielding a degraded or corrupted output (col.~4).}
    \label{fig:modality}
\end{figure}

%% file: tables/table2_threat_model.tex
\begin{table*}[t]
\centering
{\tiny
\setlength{\tabcolsep}{3pt}
\renewcommand{\arraystretch}{1.0}
\caption{\textbf{The adversarial framework of~\eqref{eq:bob_cloak}, instantiated per modality.} Each row represents one Alice--Eve--Bob scenario: Bob's cloak $\delta$, bounded by budget $\epsilon$, protects Alice's content against Eve's pipeline, entering through the structural feature map $\phi$.}
\label{tab:threat_model_norms}
\begin{tabularx}{\linewidth}{@{} l >{\raggedright\arraybackslash}X >{\raggedright\arraybackslash}X c >{\raggedright\arraybackslash}X @{}}
\toprule
\textbf{Modality} & \textbf{Eve's Pipeline (Target Task)} & \textbf{Bob's Cloak Space ($\delta$)} & \textbf{Budget ($\epsilon$)} & \textbf{Entry Point ($\phi$)} \\
\midrule
\textbf{Image} & 
\textbullet~Personalization\newline 
\textbullet~Style / copyright\newline 
\textbullet~Editing / inpainting \newline 
\textbullet~Face-swap / erasure & 
\textbullet~Additive pixel noise ($\ell_\infty$) & 
$\{4,8,16\}/255$ & 
\textbullet~VAE encoder $\mathcal{E}$ (Latent)\newline 
\textbullet~Identity map (Pixel)\newline 
\textbullet~Patch token (DiT) \\
\midrule
\textbf{Video} & 
\textbullet~I2V animation\newline 
\textbullet~Talking-head\newline 
\textbullet~V2V editing\newline 
\textbullet~T2V safety & 
\textbullet~Additive frame noise ($\ell_\infty$)\newline 
\textbullet~Optical flow shift ($\ell_2$) & 
$\{8,16,32\}/255$ & 
\textbullet~Temporal-attn. encoder\newline 
\textbullet~Spatio-temporal tokens \\
\midrule
\textbf{3D} & 
\textbullet~Content editing\newline 
\textbullet~Poisoning / comp. cost\newline 
\textbullet~Reconstruction & 
\textbullet~3D-representation params. offsets ($\Delta$)\newline 
\textbullet~Rendered-view cloak (lifted) & 
Method-specific & 
\textbullet~3D representation params. \newline 
\textbullet~Differentiable renderer $\mathcal{R}$ \\
\bottomrule
\end{tabularx}}
\end{table*}

%% file: sections/03_attacks_defenses.tex
\section{Adversarial Attacks and Defenses on Diffusion Models}
\label{sec:attacks_defenses}

Following the taxonomy of Figure~\ref{fig:hierarchy}, we organize the field by modality. Within each modality, we split methods into attacks and defenses, grouping attacks by their targeted generative task. We present attacks first, followed by defenses.
To systematically compare these approaches, our summary tables (Tables~\ref{tab:methods_image}, \ref{tab:methods_video}, and \ref{tab:methods_3d}) report the primary \textbf{evaluation benchmarks} and \textbf{target models} utilized in each work. We discuss these evaluation settings in Section~\ref{sec:evaluation}. Additionally, the tables track four critical evaluation dimensions: \textbf{Imperceptibility Analysis (IA)} (whether visual subtlety is quantitatively or qualitatively measured), \textbf{Failure Analysis (FA)} (whether edge cases and failure modes are explicitly discussed), \textbf{Transferability (TR)} (effectiveness across different model architectures or tasks), and \textbf{Robustness (RB)} (empirical testing against adaptive countermeasures like purification). In these columns, \cmark{} indicates that the paper reports the corresponding analysis, \xmark{} that it does not, and N/A that the dimension does not apply to the method (e.g., imperceptibility for prompt-level moderation, or transferability for purifiers evaluated against fixed protections).

\input{figures/fig3_task_taxonomy}

\subsection{Image}
\label{subsec:image}

\subsubsection{Attacks}
\label{subsubsec:attacks_image}

Adversarial research originated in the image domain, catalyzed by the public release of large-scale diffusion models. Because a single backbone supports numerous downstream tasks, each acts as a unique attack surface. We categorize these methods below by their targeted generative task, structured chronologically within Table~\ref{tab:methods_image}.

\paragraph{Personalization.}
\label{subsubsec:attacks_personalization}
Personalization methods fine-tune a diffusion model on a few images of a subject so it can regenerate that subject in new contexts; misused, they enable identity theft and non-consensual imagery, so the goal of a cloak here is to make fine-tuning on the protected images fail. \textbf{Anti-DreamBooth}~\cite{van2023anti} set the template and remains the baseline: it optimizes $\delta$ to corrupt the DreamBooth personalization loop itself, instantiating the bilevel objective of~\eqref{eq:ft}. \textbf{SimAC}~\cite{wang2024simac} makes anti-customization stronger by attacking the useful timesteps and features, \textbf{InMark}~\cite{liu2024countering} embeds protective watermarks on influential pixels, using influence-function-inspired pixel selection so that personalized fine-tuning fails even after common image modifications, \textbf{MetaCloak}~\cite{liu2024metacloak} learns transferable, robust perturbations via meta-learning, transfers cloak to a commercial service (Replicate), \textbf{DisDiff}~\cite{liu2024disrupting} erases subject-token attention to break the text-subject binding, and \textbf{DADiff}~\cite{tang2025make} combines prompt-level and image-level adversarial attacks to disrupt customization more thoroughly. A later line hardens the cloak against purification: \textbf{HF-Anti-DreamBooth}~\cite{onikubo2024hf} concentrates perturbation power in high-frequency regions of the image that low-pass purifiers miss, \textbf{AntiPure}~\cite{yang2025antipure} constructs perturbations that persist through diffusion-based purification, and \textbf{LDU}~\cite{devulapally2025latent} shifts the denoising trajectory of diffusion models. \textbf{FastProtect}~\cite{ahn2024nearly} reduces cost by replacing per-image optimization with a pre-trained perturbation generator (200--3500$\times$ faster). \textbf{IDDM}~\cite{dai2026iddm} protects personalized generation while disrupting face recognition of the generated images. \textbf{VCPro}~\cite{mi2024visual} confines the least-perceptible cloak to user-masked key concepts.

\afterpage{\clearpage\input{tables/table3_methods_image}}

\paragraph{Style\,/\,copyright.}
\label{subsubsec:attacks_style}
Style mimicry fine-tunes a model on an artist's works so it can reproduce their signature style on demand, threatening artists' livelihoods and copyright; a style cloak aims to make the model learn the wrong style from the protected artworks. \textbf{Glaze}~\cite{glaze} brought cloaking to working artists: its style cloaks make models fine-tuned on the protected artworks learn the wrong artistic style (an end-user tool protecting over 2M images), and it first flagged purification as a threat. \textbf{Mist}~\cite{liang2023mist} increases transferability with a fused adversarial loss term. Later work diversified whose identity is protected and where the cloak acts: \textbf{ID-Cloak}~\cite{teng2025id} learns identity-specific universal cloaks by modeling an identity subspace and optimizing a cloak that transfers across images of the same person, \textbf{SITA}~\cite{kang2025sita} uses CLIP to decouple and disrupt the style representation of the image when the fine-tuning method is unknown, and \textbf{StyleProtect}~\cite{tang2025styleprotect} targets selected cross-attention layers.

\paragraph{Editing\,/\,inpainting.}
\label{subsubsec:attacks_editing}
\label{subsubsec:attacks_inpainting}
Text-guided editing takes a real image and a text instruction and returns an altered version of it, while inpainting fills a masked region conditioned on the surrounding content and a prompt; both enable unauthorized manipulation of published photos, so the goal of a cloak here is to derail the edit and leave the output unusable. \textbf{PhotoGuard}~\cite{salman2023raising} set the template and remains the de-facto baseline: it offers a cheap VAE-encoder attack and a stronger end-to-end diffusion attack that steers generation toward a target image. \textbf{AdvDM}~\cite{advdm} formulates protection as minimizing the diffusion-model likelihood of a perturbed image, then optimizes a tractable surrogate by maximizing the expected diffusion loss over sampled latent trajectories. \textbf{SDS}~\cite{sdsattack} accelerates diffusion-loss attacks via score distillation sampling. \textbf{EditShield}~\cite{chen2024editshield} protects against instruction-guided diffusion editors such as InstructPix2Pix~\cite{instructpix2pix} by adding imperceptible perturbations that shift the instruction-independent conditioning latent, and \textbf{Anti-Reference}~\cite{antireference} uses a unified weighted loss to jointly target tuning-based customization methods and reference-conditioned generation modules. \textbf{TarPro}~\cite{shen2026tarpro} uses a semantic-aware constraint to suppress malicious prompt components while preserving the benign parts of the edit, \textbf{ACE}~\cite{zheng2025targeted} makes the attack targeted, steering the predicted score toward one fixed chaotic pattern. \textbf{FaceLock}~\cite{wang2024facelock} erases the subject's biometric information,  \textbf{AdvI2I}~\cite{zeng2024advi2i} trains a generator whose images drive image-to-image models toward unsafe outputs under benign prompts, and \textbf{Anti-Diffusion}~\cite{zheng2025antidiffusion} adds a semantic disturbance loss that drives cross-attention maps toward a zero target. A parallel line hardens cloaks against compression and purification: \textbf{DCT-Shield}~\cite{dct-shield} optimizes perturbations over quantized DCT coefficients produced by a JPEG encoding--decoding pipeline, and \textbf{BlurGuard}~\cite{kim2025blurguard} adaptively blurs the cloak per semantic region so the protected image better preserves the original image's power spectrum. As backbones diversified, cloaks followed: \textbf{PCA}~\cite{guo2025pca} collapses the VAE posterior via KL manipulation, \textbf{AtkPDM}~\cite{xue2024atkpdm} attacks intermediate U-Net features of pixel-domain models, and \textbf{DeContext}~\cite{shen2025decontext} disrupts the in-context cross-attention flow of DiT editors (FLUX.1 Kontext \cite{labs2025flux}). On the inpainting side, \textbf{AdvPaint}~\cite{jeon2025advpaint} disrupts self- and cross-attention with separate perturbations inside and outside an enlarged object box, \textbf{DiffusionGuard}~\cite{choi2024diffusionguard} disrupts the early stages of the diffusion process and hardens the cloak via mask augmentation, and introduces InpaintGuardBench, \textbf{PromptFlare}~\cite{na2025promptflare} drives masked-region cross-attention toward the prompt-invariant BOS token, \textbf{SDA}~\cite{he2025structure} perturbs first-step self-attention queries to break early contour formation, and \textbf{Anti-Inpainting}~\cite{guo2025antiinpainting} extends protection to unknown masks, prompts, and seeds. Finally, \textbf{Universal Immunization}~\cite{lee2026universal} amortizes the attack with a single image-agnostic perturbation, and \textbf{DiffVax}~\cite{ozden2026diffvax} trains a feed-forward immunizer that cloaks unseen images in one pass ($\sim$250{,}000$\times$ faster), extending to video.

\paragraph{Face-swap\,/\,biometric.}
\label{subsubsec:attacks_faceswap}
Face swapping transplants a target identity onto another face, the core mechanism behind deepfakes; cloaks here aim to make swappers extract the wrong identity. \textbf{FaceSwapGuard}~\cite{wang2025faceswapguard} adds imperceptible perturbations that disrupt identity-feature extraction in black-box face-swapping models; \textbf{My Face Is Mine}~\cite{yam2025myfaceismine} uses identity and timestep-averaged deviation losses to craft memory-efficient LDM-latent perturbations that transfer across diffusion-based face swappers; \textbf{FaceShield}~\cite{jeong2024faceshield} disrupts diffusion face-swappers via source-conditioning attention attacks, facial-feature-extractor attacks, and blur/low-pass noise updates for imperceptibility and JPEG robustness; and \textbf{AEGIS}~\cite{li2026aegis} injects adversarial signals into the DDIM denoising trajectory, avoiding the pixel-space peak-clipping constraint and disrupting both GAN- and diffusion-based facial manipulation.

\subsubsection{Defenses}
\label{subsubsec:defenses_image}
Here ``defense'' spans two arenas: (1)~\emph{purification}, Eve's counters that strip a cloak so her pipeline runs as intended, and (2)~\emph{unlearning and red-teaming}, where a cooperating provider removes or blocks the exploitable capability in the model itself, and red-teamers probe how durable that removal is.
\paragraph{Purification.} The direct countermeasure is to preprocess the protected image $x{+}\delta$ before editing or customization, and even simple lossy transformations strip many cloaks: JPEG compression substantially weakens \textbf{PhotoGuard}-style perturbations~\cite{sandoval2023jpeg}, and black-box \textbf{Noisy Upscaling}~\cite{honig2024adversarial}---injecting noise, then reconstructing the image with a generative model---bypasses \textbf{Glaze}, \textbf{Mist}, and \textbf{Anti-DreamBooth}. \textbf{DiffPure}~\cite{diffpure} provides the canonical diffusion-purification formulation, smoothing out adversarial perturbations while preserving image semantics; \textbf{GrIDPure}~\cite{gridpure} adapts it to high-resolution images via small-step purification on overlapping grids with patch averaging; and \textbf{PDM-Pure}~\cite{xue2024pixel}, arguing that perturbations crafted against LDMs transfer poorly to pixel-space diffusion models, purifies with a PDM instead. Since Eve can further modify the customization pipeline itself, many protective perturbations resist neither input-level restoration nor model-level adaptation---motivating newer cloaks that explicitly optimize for survival under purification.
\paragraph{Unlearning and red-teaming.} The lightest model-side interventions act at inference: \textbf{SLD}~\cite{schramowski2023safe} adds a safety term to classifier-free guidance that steers denoising away from a text-defined unsafe direction without retraining, and \textbf{GuardT2I}~\cite{yang2024guardt2i} translates guidance embeddings back into natural language to expose intent hidden by adversarial prompts. Training-time erasure instead scrubs the capability from the weights: \textbf{EraseDiff}~\cite{wu2024erasediff} redirects forget-set denoising toward mismatched targets for fast unlearning, and \textbf{Erasing}~\cite{fuchi2024erasing} fine-tunes only the CLIP text encoder on a few images in seconds, though broad concepts resist. \textbf{GuardDoor}~\cite{zeng2025guarddoor} sidesteps the purification arms race by baking a trigger-activated backdoor into the image encoder that survives JPEG, \textbf{DiffPure}, and IMPRESS, at the cost of leaving non-participating editors untouched. Yet automated red-teaming that optimizes adversarial prompts against a guarded model routinely revives ``erased'' concepts, exposing how shallow such safety can be; \textbf{AdvUnlearn}~\cite{zhang2024advunlearn} responds by adversarially training the CLIP text encoder with utility-preserving regularization, transferring modularly across diffusion backbones, though continuous textual-inversion attacks remain a harder failure mode. Red-teaming also targets the cloaks themselves: \textbf{DiffShortcut}~\cite{liu2024rethinkred} purifies protected inputs with restoration and super-resolution, then disentangles the personalized concept from residual noise via contrastive decoupling learning, while \textbf{CAT}~\cite{peng2025catadvtrain} trains lightweight LoRA adapters in the LDM autoencoder to realign latent representations distorted by protected samples, without purifying pixels.

\subsection{Video}
\label{subsec:video}

\input{tables/table4_methods_video}

\subsubsection{Attacks} Video is the fastest-growing surface, contending with the temporal axis and aggressively lossy codecs.
\label{subsubsec:attacks_video}
\paragraph{I2V.}
\label{subsubsec:image2video}
I2V pipelines (ControlNeXt~\cite{peng_controlnext_2025}, SVD \cite{blattmann2023stable}, CogVideoX~\cite{yang_cogvideox_2024}) animate a reference image, risking misleading content. \textbf{I2VGuard}~\cite{gui_i2vguard_nodate}, the first white-box I2V defense, composes spatial, temporal, and contrastive-latent losses; \textbf{UVCG}~\cite{li2024uvcg} makes one perturbation universal across videos; \textbf{Vid-Freeze}~\cite{chowdhury2026vidfreeze} freezes motion to the first frame (plausible but static); and \textbf{Anti-I2V}~\cite{vu2026antii2v} handles DiT-based I2V via L*a*b* color space / frequency domain-collapse losses, since U-Net cloaks fail on DiTs. \textbf{DORMANT}~\cite{zhou2025dormant} extends protection to pose-driven human-image animation, misextracting CLIP and ReferenceNet features with EoT and momentum. \textbf{Immune2V}~\cite{long2026immune2v}, the most principled I2V defense, explains why image cloaks fail on dual-stream I2V (per-frame noise \emph{dilutes} and text guidance \emph{overrides} them) and counters with a temporally balanced latent-divergence loss plus alignment to a collapse-inducing trajectory, yielding stronger, more persistent degradation.

\paragraph{Talking-head.}
\label{subsubsec:talkinghead}
\textbf{Silencer}~\cite{gan2025silencer} nullifies audio control (keeping the mouth closed), then aligns DDIM latents for anti-purification; \textbf{SyncBreaker}~\cite{zhang2026syncbreaker} attacks the image and audio streams with stage-awareness.

\paragraph{V2V editing.}
\label{subsubsec:video2video}
In video-to-video (\vtov) editing, \textbf{PRIME}~\cite{li2024prime} runs per-frame \textbf{PhotoGuard} with in-loop codec simulation at ${\sim}90\%$ lower cost; \textbf{VideoGuard}~\cite{cao_videoguard_2025} then tackles codec non-differentiability directly, optimizing a DDIM latent before gradient-free PSO on pixels; and \textbf{VGMShield}~\cite{pang2025vgmshield} broadens the scope to lifecycle detection, attribution, and prevention.

\paragraph{T2V safety.}
\label{subsubsec:t2v_attacks}
\textbf{T2VAttack}~\cite{li2025t2vattack} perturbs prompts for semantic/temporal failure and \textbf{T2V-OptJail}~\cite{liu2025t2voptjail} jailbreaks commercial T2V (${\sim}64\%$); in response, \textbf{T2VShield}~\cite{liang2025t2vshield} rewrites prompts and detects unsafe output (${\sim}33\%$ ASR drop) and \textbf{ConceptGuard}~\cite{ma2025conceptguard} flags TI2V risk (ConceptRisk), both evaluated by T2VSafetyBench~\cite{miao2024t2vsafetybench}.

\subsubsection{Defenses}
\label{subsubsec:defenses_video}

Counter-defenses against video cloaks are still nascent: existing work treats preprocessing and purification as robustness stress tests rather than dedicated removal methods, in contrast to image-domain purification and adaptive counterattacks. \textbf{IPV-Bench}~\cite{li2026ipbench} provides the clearest evidence, evaluating I2V protection under preprocessing attacks and transfer settings and showing that existing protections remain fragile to practical transformations.


\input{tables/table5_methods_3d}

\subsection{3D}
\label{subsec:3d}

\subsubsection{Attacks} 3D representations are not diffusion models themselves, but they have become an emerging security surface because many 3D assets are now edited through rendered views and diffusion-based guidance.
\label{subsubsec:attacks_3d}

\paragraph{Editing.} \textbf{DEGauss}~\cite{meng2025degauss} protects Gaussian splatting assets by optimizing perturbations over Gaussian representations with view-focal gradient fusion and dual-discrepancy objectives, while \textbf{AdLift}~\cite{hong2025adlift} lifts bounded rendered-view perturbations into safeguard Gaussians so that the protection remains consistent under novel-view interpolation rather than being smoothed away like a naive 2D cloak. Beyond \gsplat editing, \textbf{Image2Multiview}~\cite{sun2024latent} protects 3D assets by perturbing the source image to disrupt multi-view diffusion and downstream reconstruction.

\paragraph{Poisoning\,/\,cost.} Recent work also studies broader \gsplat attack surfaces. On the poisoning side, \textbf{StealthAttack}~\cite{ke2025stealthattack} injects density-guided Gaussian illusions into low-density regions so that malicious objects appear only from selected viewpoints, and \textbf{GaussTrap}~\cite{hong2025gausstrap} studies stealthy poisoning/backdoor attacks that cause targeted scene confusion while preserving benign views. At the system level, \textbf{Poison-splat}~\cite{lu2025poison} attacks the computational cost of \gsplat reconstruction. 
Orthogonal to cloaking, \textbf{RDSplat}~\cite{zhao2025rdsplat} studies watermarking that remains robust under diffusion-based \gsplat editing.

\subsubsection{Defenses}
\label{subsubsec:defenses_3d}
Counter-defenses for 3D content remain limited. Existing defenses mainly cover adjacent 3DGS threat models: \textbf{RemedyGS}~\cite{li2025remedygs} uses detect-and-purify mechanisms to protect reconstruction pipelines from poisoned inputs that inflate computation cost, while \textbf{RDSplat}~\cite{zhao2025rdsplat} focuses on provenance rather than cloak removal by embedding watermarks robust to diffusion-based 3DGS editing. These works indicate early system-level and provenance-oriented protections, while general counter-defenses for 3D content cloaks remain open.

%% file: figures/fig3_task_taxonomy.tex
\providecommand{\taxleaf}[1]{\parbox{7.0cm}{\raggedright #1}}
\providecommand{\taxleafS}[1]{\parbox{3.3cm}{\raggedright #1}}
\providecommand{\taxtaskB}[1]{\parbox{1.95cm}{\centering #1}}
\providecommand{\taxtaskS}[1]{\parbox{1.5cm}{\centering #1}}
\providecommand{\taxsplit}[1]{\parbox{1.2cm}{\centering #1}}
\begin{figure*}[!t]
\centering
\begin{subfigure}[b]{\linewidth}
\centering
\adjustbox{max width=\linewidth, max totalheight=0.40\textheight}{%
\begin{forest}
  forked edges,
  for tree={font=\sffamily,grow'=east,parent anchor=east,child anchor=west,anchor=west,calign=first,
    edge={line width=0.5pt},rounded corners=2pt,inner xsep=3pt,inner ysep=1.6pt,l sep=2mm,s sep=0.4mm,fork sep=1.6mm},
  Si/.style={font=\sffamily\bfseries\tiny,align=center,anchor=center,calign=center,rounded corners=3pt,fill=blue!38,draw=blue!60},
  Ti/.style={font=\sffamily\bfseries\tiny,fill=blue!28,draw=blue!50},
  Li/.style={font=\sffamily\tiny,rounded corners=2pt,fill=blue!11,draw=blue!35},
  defS/.style={font=\sffamily\bfseries\tiny,align=center,anchor=center,calign=center,rounded corners=3pt,fill=defcol!30,draw=defcol!62},
  defT/.style={font=\sffamily\bfseries\tiny,fill=defcol!22,draw=defcol!60},
  defL/.style={font=\sffamily\tiny,rounded corners=2pt,fill=defcol!14,draw=defcol!55},
[, phantom, for children={no edge}
  [{\taxsplit{Attacks}}, Si
    [{\taxtaskB{Personalization}}, Ti [{\taxleaf{\textcolor{atkcol}{Anti-DreamBooth~\cite{van2023anti}}, \textcolor{atkcol}{SimAC~\cite{wang2024simac}}, \textcolor{atkcol}{InMark~\cite{liu2024countering}}, \textcolor{atkcol}{MetaCloak~\cite{liu2024metacloak}}, \textcolor{atkcol}{HF-Anti-DreamBooth~\cite{onikubo2024hf}}, \textcolor{atkcol}{DisDiff~\cite{liu2024disrupting}}, \textcolor{atkcol}{FastProtect~\cite{ahn2024nearly}}, \textcolor{atkcol}{AntiPure~\cite{yang2025antipure}}, \textcolor{atkcol}{LDU~\cite{devulapally2025latent}}, \textcolor{atkcol}{DADiff~\cite{tang2025make}}, \textcolor{atkcol}{IDDM~\cite{dai2026iddm}}, \textcolor{atkcol}{VCPro~\cite{mi2024visual}}}}, Li]]
    [{\taxtaskB{Style / copyright}}, Ti [{\taxleaf{\textcolor{atkcol}{Glaze~\cite{glaze}}, \textcolor{atkcol}{Mist~\cite{liang2023mist}}, \textcolor{atkcol}{ID-Cloak~\cite{teng2025id}}, \textcolor{atkcol}{SITA~\cite{kang2025sita}}, \textcolor{atkcol}{StyleProtect~\cite{tang2025styleprotect}}}}, Li]]
    [{\taxtaskB{Editing / Inpainting}}, Ti [{\taxleaf{\textcolor{atkcol}{PhotoGuard~\cite{salman2023raising}}, \textcolor{atkcol}{AdvDM~\cite{advdm}}, \textcolor{atkcol}{SDS~\cite{sdsattack}}, \textcolor{atkcol}{EditShield~\cite{chen2024editshield}}, \textcolor{atkcol}{Anti-Reference~\cite{antireference}}, \textcolor{atkcol}{Anti-Diffusion~\cite{zheng2025antidiffusion}}, \textcolor{atkcol}{AtkPDM~\cite{xue2024atkpdm}}, \textcolor{atkcol}{TarPro~\cite{shen2026tarpro}}, \textcolor{atkcol}{ACE~\cite{zheng2025targeted}}, \textcolor{atkcol}{AdvPaint~\cite{jeon2025advpaint}}, \textcolor{atkcol}{DiffusionGuard~\cite{choi2024diffusionguard}}, \textcolor{atkcol}{FaceLock~\cite{wang2024facelock}}, \textcolor{atkcol}{AdvI2I~\cite{zeng2024advi2i}}, \textcolor{atkcol}{DCT-Shield~\cite{dct-shield}}, \textcolor{atkcol}{PromptFlare~\cite{na2025promptflare}}, \textcolor{atkcol}{BlurGuard~\cite{kim2025blurguard}}, \textcolor{atkcol}{PCA~\cite{guo2025pca}}, \textcolor{atkcol}{DeContext~\cite{shen2025decontext}}, \textcolor{atkcol}{SDA~\cite{he2025structure}}, \textcolor{atkcol}{Anti-Inpainting~\cite{guo2025antiinpainting}}, \textcolor{atkcol}{DiffVax~\cite{ozden2026diffvax}}, \textcolor{atkcol}{Universal Immunization~\cite{lee2026universal}}}}, Li]]
    [{\taxtaskB{Face-swap / biometric}}, Ti [{\taxleaf{\textcolor{atkcol}{FaceShield~\cite{jeong2024faceshield}}, \textcolor{atkcol}{FaceSwapGuard~\cite{wang2025faceswapguard}}, \textcolor{atkcol}{My Face Is Mine~\cite{yam2025myfaceismine}}, \textcolor{atkcol}{AEGIS~\cite{li2026aegis}}}}, Li]]
  ]
  [{\taxsplit{Defenses}}, defS
    [{\taxtaskB{Purification}}, defT [{\taxleaf{\textcolor{defcol}{DiffPure~\cite{diffpure}}, \textcolor{defcol}{GrIDPure~\cite{gridpure}}, \textcolor{defcol}{PDM-Pure~\cite{xue2024pixel}}, \textcolor{defcol}{Noisy Upscaling~\cite{honig2024adversarial}}}}, defL]]
    [{\taxtaskB{Unlearning, Red-Teaming}}, defT [{\taxleaf{\textcolor{defcol}{SLD~\cite{schramowski2023safe}}, \textcolor{defcol}{AdvUnlearn~\cite{zhang2024advunlearn}}, \textcolor{defcol}{GuardT2I~\cite{yang2024guardt2i}}, \textcolor{defcol}{EraseDiff~\cite{wu2024erasediff}}, \textcolor{defcol}{Erasing~\cite{fuchi2024erasing}}, \textcolor{defcol}{CAT~\cite{peng2025catadvtrain}}, \textcolor{defcol}{GuardDoor~\cite{zeng2025guarddoor}}, \textcolor{defcol}{DiffShortcut~\cite{liu2024rethinkred}}}}, defL]]
  ]
]
\end{forest}}%
\caption{Image}
\label{fig:hierarchy_image}
\end{subfigure}
\vspace{0.7ex}
\begin{subfigure}[b]{0.58\linewidth}
\centering
\adjustbox{max width=\linewidth, max totalheight=0.30\textheight}{%
\begin{forest}
  forked edges,
  for tree={font=\sffamily,grow'=east,parent anchor=east,child anchor=west,anchor=west,calign=first,
    edge={line width=0.5pt},rounded corners=2pt,inner xsep=3pt,inner ysep=1.6pt,l sep=2mm,s sep=0.4mm,fork sep=1.6mm},
  Sv/.style={font=\sffamily\bfseries\tiny,align=center,anchor=center,calign=center,rounded corners=3pt,fill=red!38,draw=red!60},
  Tv/.style={font=\sffamily\bfseries\tiny,fill=red!28,draw=red!50},
  Lv/.style={font=\sffamily\tiny,rounded corners=2pt,fill=red!11,draw=red!35},
[, phantom, for children={no edge}
  [{\taxsplit{Attacks}}, Sv
    [{\taxtaskS{I2V}}, Tv [{\taxleafS{\textcolor{atkcol}{UVCG~\cite{li2024uvcg}}, \textcolor{atkcol}{I2VGuard~\cite{gui_i2vguard_nodate}}, \textcolor{atkcol}{DORMANT~\cite{zhou2025dormant}}, \textcolor{atkcol}{Vid-Freeze~\cite{chowdhury2026vidfreeze}}, \textcolor{atkcol}{Anti-I2V~\cite{vu2026antii2v}}, \textcolor{atkcol}{Immune2V~\cite{long2026immune2v}}}}, Lv]]
    [{\taxtaskS{Talking-head}}, Tv [{\taxleafS{\textcolor{atkcol}{Silencer~\cite{gan2025silencer}}, \textcolor{atkcol}{SyncBreaker~\cite{zhang2026syncbreaker}}}}, Lv]]
    [{\taxtaskS{V2V editing}}, Tv [{\taxleafS{\textcolor{atkcol}{PRIME~\cite{li2024prime}}, \textcolor{atkcol}{VGMShield~\cite{pang2025vgmshield}}, \textcolor{atkcol}{VideoGuard~\cite{cao_videoguard_2025}}}}, Lv]]
    [{\taxtaskS{T2V safety}}, Tv [{\taxleafS{\textcolor{atkcol}{T2VAttack~\cite{li2025t2vattack}}, \textcolor{atkcol}{T2V-OptJail~\cite{liu2025t2voptjail}}, \textcolor{atkcol}{ConceptGuard~\cite{ma2025conceptguard}}, \textcolor{atkcol}{T2VShield~\cite{liang2025t2vshield}}}}, Lv]]
  ]
]
\end{forest}}%
\caption{Video}
\label{fig:hierarchy_video}
\end{subfigure}%
\hfill
\begin{subfigure}[b]{0.40\linewidth}
\centering
\adjustbox{max width=\linewidth, max totalheight=0.30\textheight}{%
\begin{forest}
  forked edges,
  for tree={font=\sffamily,grow=south,parent anchor=south,child anchor=north,anchor=north,calign=center,
    edge={line width=0.5pt},rounded corners=2pt,inner xsep=3pt,inner ysep=1.6pt,l sep=2.5mm,s sep=1.5mm,fork sep=1.4mm},
  Sg/.style={font=\sffamily\bfseries\tiny,align=center,rounded corners=3pt,fill=violet!38,draw=violet!60},
  Tg/.style={font=\sffamily\bfseries\tiny,align=center,fill=violet!28,draw=violet!50},
  Lg/.style={font=\sffamily\tiny,rounded corners=2pt,fill=violet!11,draw=violet!35},
  defS/.style={font=\sffamily\bfseries\tiny,align=center,rounded corners=3pt,fill=defcol!30,draw=defcol!62},
  defL/.style={font=\sffamily\tiny,rounded corners=2pt,fill=defcol!14,draw=defcol!55},
[, phantom, for children={no edge}
  [{\taxsplit{Attacks}}, Sg
    [{Editing}, Tg [{\parbox{2.1cm}{\centering\textcolor{atkcol}{Image2Multiview~\cite{sun2024latent}}\\[1pt]\textcolor{atkcol}{DEGauss~\cite{meng2025degauss}}\\[1pt]\textcolor{atkcol}{AdLift~\cite{hong2025adlift}}}}, Lg]]
    [{Poisoning / cost}, Tg [{\parbox{2.1cm}{\centering\textcolor{atkcol}{Poison-splat~\cite{lu2025poison}}\\[1pt]\textcolor{atkcol}{StealthAttack~\cite{ke2025stealthattack}}\\[1pt]\textcolor{atkcol}{GaussTrap~\cite{hong2025gausstrap}}}}, Lg]]
  ]
  [{\taxsplit{Defenses}}, defS [{\parbox{1.9cm}{\centering\textcolor{defcol}{RDSplat~\cite{zhao2025rdsplat}}\\[1pt]\textcolor{defcol}{RemedyGS~\cite{li2025remedygs}}}}, defL]]
]
\end{forest}}%
\caption{3D}
\label{fig:hierarchy_3d}
\end{subfigure}
\caption{\textbf{Task-centric taxonomy of adversarial attacks and defenses for diffusion models across (a)~image, (b)~video, and (c)~3D.} Methods are grouped by generative task within each modality; image and 3D defenses are collected in separate branches, while video counter-defenses remain nascent and are discussed in the text (\S\ref{subsec:video}). \textcolor{atkcol}{\textbf{Red}} marks attacks and \textcolor{defcol}{\textbf{green}} marks defenses.}
\label{fig:hierarchy}
\end{figure*}
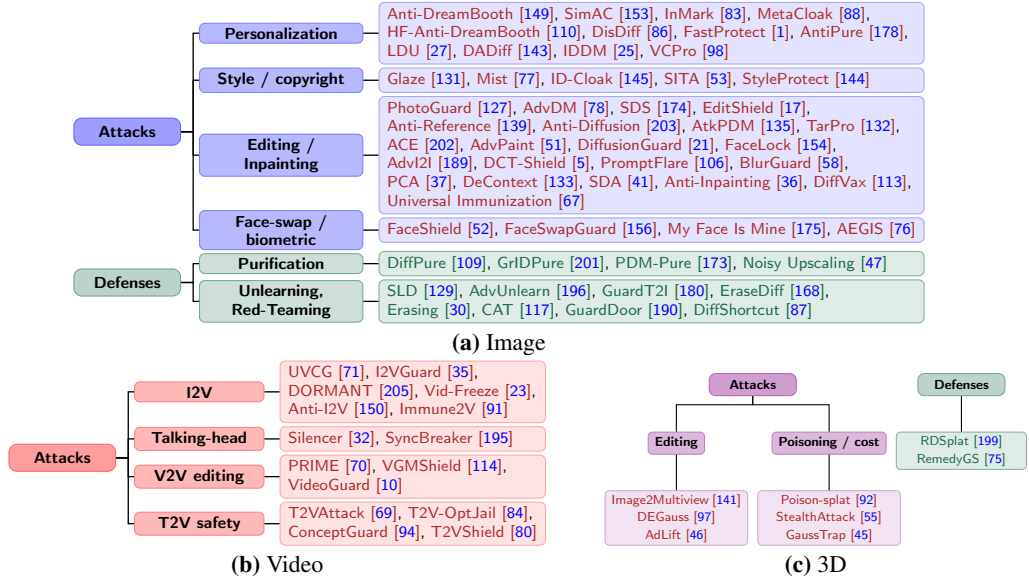

%% file: tables/table3_methods_image.tex
\clearpage
\definecolor{defcol}{RGB}{30,110,75}
\colorlet{bandDef}{defcol!22}
\colorlet{rowshade}{gray!13}
{\fontsize{5.0pt}{5.7pt}\selectfont
\setlength{\tabcolsep}{2.2pt}
\renewcommand{\arraystretch}{0.95}
\setlength{\LTleft}{\fill}\setlength{\LTright}{\fill}
\setlength{\LTpre}{4pt}\setlength{\LTpost}{4pt}
\setlength{\tblrest}{\dimexpr\linewidth-1.50cm-1.36cm-14\tabcolsep\relax}
\begin{longtable}{@{}
>{\rowstrut\raggedright\arraybackslash}m{0.22\tblrest}
>{\centering\arraybackslash}m{1.50cm}
>{\raggedright\arraybackslash}m{0.45\tblrest}
>{\raggedright\arraybackslash}m{0.33\tblrest}
>{\centering\arraybackslash}m{0.34cm}
>{\centering\arraybackslash}m{0.34cm}
>{\centering\arraybackslash}m{0.34cm}
>{\centering\arraybackslash}m{0.34cm}
@{}}
\caption{\textbf{Image attack and defense methods.}}\label{tab:methods_image}\\[1pt]
\toprule
\textbf{Method} & \textbf{Venue} & \textbf{Evaluation Sources}
& \textbf{Target Models}
& \textbf{IA} & \textbf{FA} & \textbf{TR} & \textbf{RB} \\
\midrule
\endfirsthead
\multicolumn{8}{@{}l}{\emph{Table~\ref{tab:methods_image} continued.}} \\
\toprule
\textbf{Method} & \textbf{Venue} & \textbf{Evaluation Sources}
& \textbf{Target Models}
& \textbf{IA} & \textbf{FA} & \textbf{TR} & \textbf{RB} \\
\midrule
\endhead
\midrule
\multicolumn{8}{r@{}}{\emph{(continued)}} \\
\endfoot
\bottomrule
\endlastfoot
\grouprowImg{bandImg}{Personalization}
Anti-DreamBooth~\cite{van2023anti} & ICCV'23 & CelebA-HQ, VGGFace2 & DreamBooth, TI, DreamBooth w/ LoRA & \cmark & \xmark & \cmark & \cmark \\
\rowcolor{rowshade}
SimAC~\cite{wang2024simac} & CVPR'24 & CelebA-HQ, VGGFace2 & DreamBooth, LoRA, Custom Diffusion & \cmark & \xmark & \cmark & \xmark \\
InMark~\cite{liu2024countering} & CVPR'24 & VGGFace2, WikiArt & DreamBooth, TI, LoRA & \cmark & \xmark & \cmark & \cmark \\
\rowcolor{rowshade}
MetaCloak~\cite{liu2024metacloak} & CVPR'24 & CelebA-HQ, VGGFace2 & DreamBooth, Replicate & \xmark & \xmark & \cmark & \cmark \\
HF-Anti-DreamBooth~\cite{onikubo2024hf} & ECCVW'24 & VGGFace2 & DreamBooth & \xmark & \xmark & \xmark & \cmark \\
\rowcolor{rowshade}
DisDiff~\cite{liu2024disrupting} & ACM~MM'24 & CelebA-HQ, VGGFace2 & DreamBooth, LoRA, TI & \xmark & \xmark & \cmark & \xmark \\
FastProtect~\cite{ahn2024nearly} & CVPR'25 & ImageNet, FFHQ, WikiArt, WebToon & LoRA, SD2.1, SDXL, TI, DreamStyler & \cmark & \cmark & \cmark & \cmark \\
\rowcolor{rowshade}
AntiPure~\cite{yang2025antipure} & ICCV'25 & CelebA-HQ, VGGFace2 & DreamBooth, LoRA & \cmark & \xmark & \xmark & \cmark \\
LDU~\cite{devulapally2025latent} & ACM~MM'25 & CelebA-HQ, VGGFace2, WikiArt & DreamBooth, TI, SD1.5, SD2.1 & \cmark & \xmark & \cmark & \cmark \\
\rowcolor{rowshade}
DADiff~\cite{tang2025make} & arXiv'25 & CelebA-HQ, VGGFace2 & DreamBooth & \xmark & \xmark & \cmark & \xmark \\
IDDM~\cite{dai2026iddm} & arXiv'26 & CelebA-HQ, VGGFace2 & DreamBooth, LoRA & \cmark & \xmark & \cmark & \cmark \\
\rowcolor{rowshade}
VCPro~\cite{mi2024visual} & AAAI'26 & CelebA-HQ, VGGFace2 & SD1.4 & \cmark & \cmark & \cmark & \cmark \\
\grouprowImg{bandImg}{Style / copyright}
Glaze~\cite{glaze} & USENIX'23 & Current artists, WikiArt & Stable Diffusion, DALL$\cdot$E-mega & \cmark & \cmark & \cmark & \cmark \\
\rowcolor{rowshade}
Mist~\cite{liang2023mist} & arXiv'23 & WikiArt & Stable Diffusion, NovelAI & \cmark & \xmark & \cmark & \cmark \\
ID-Cloak~\cite{teng2025id} & arXiv'25 & CelebA-HQ, VGGFace2 & SD1.5, SD2.1, DreamBooth, DreamBooth-LoRA, TI & \xmark & \xmark & \cmark & \xmark \\
\rowcolor{rowshade}
SITA~\cite{kang2025sita} & TIFS'25 & WikiArt, ArtBench & T2I-Adapter, TI, DreamBooth & \cmark & \cmark & \cmark & \cmark \\
StyleProtect~\cite{tang2025styleprotect} & CVPR~W'26 & WikiArt, Anita & SD1.5 & \cmark & \xmark & \xmark & \cmark \\
\grouprowImg{bandImg}{Editing / Inpainting}
PhotoGuard~\cite{salman2023raising} & ICML'23 & 60 Curated Images & SD1.5 & \xmark & \cmark & \xmark & \xmark \\
\rowcolor{rowshade}
AdvDM~\cite{advdm} & ICML'23 & LSUN, WikiArt & LDM & \cmark & \xmark & \cmark & \cmark \\
SDS~\cite{sdsattack} & ICLR'24 & Crawled landscape, anime, and portrait pictures; WikiArt & SD LDM & \cmark & \cmark & \cmark & \cmark \\
\rowcolor{rowshade}
EditShield~\cite{chen2024editshield} & ECCV'24 & InstructPix2Pix, MagicBrush & ip2p/ip2p-mb & \xmark & \cmark & \cmark & \cmark \\
Anti-Reference~\cite{antireference} & arXiv'24 & DreamBooth, CelebA-HQ, TikTok & DreamBooth, LoRA, TI, IP-Adapter, Reference-only, Magic Animate, EchoMimic & \cmark & \cmark & \cmark & \cmark \\
\rowcolor{rowshade}
Anti-Diffusion~\cite{zheng2025antidiffusion} & AAAI'25 & VGGFace2, CelebA-HQ, Defense-Edit & DreamBooth, LoRA, MasaCtrl, DiffEdit, SD2.1 & \cmark & \xmark & \cmark & \xmark \\
AtkPDM~\cite{xue2024atkpdm} & AAAI'25 & 30 images per PDM (half training data, half web-crawled) & Unconditional pixel-domain DDPMs & \cmark & \xmark & \cmark & \cmark \\
\rowcolor{rowshade}
TarPro~\cite{shen2026tarpro} & AAAI'25 & Synthetic individuals from Midjourney's gallery & InstructPix2Pix, MagicBrush, HQ-Edit & \cmark & \xmark & \xmark & \xmark \\
ACE~\cite{zheng2025targeted} & ICLR'25 & CelebA-HQ, WikiArt & LoRA+DreamBooth, SDEdit & \cmark & \cmark & \cmark & \cmark \\
\rowcolor{rowshade}
AdvPaint~\cite{jeon2025advpaint} & ICLR'25 & 100 images from Pexels, Unsplash & SD Inpainting (SD1.2) & \cmark & \cmark & \cmark & \cmark \\
DiffusionGuard~\cite{choi2024diffusionguard} & ICLR'25 & InpaintGuardBench & SD Inpainting & \cmark & \xmark & \cmark & \cmark \\
\rowcolor{rowshade}
FaceLock~\cite{wang2024facelock} & CVPR'25 & CelebA-HQ & ip2p & \cmark & \xmark & \xmark & \cmark \\
AdvI2I~\cite{zeng2024advi2i} & ICML'25 & 2k NSFW pairs & ip2p/SD1.5 & \cmark & \cmark & \cmark & \cmark \\
\rowcolor{rowshade}
DCT-Shield~\cite{dct-shield} & ICCV'25 & OmniEdit, PPR10K & ip2p/SD1.0 & \cmark & \cmark & \cmark & \cmark \\
PromptFlare~\cite{na2025promptflare} & ACM~MM'25 & EditBench & SD Inpainting & \cmark & \cmark & \cmark & \cmark \\
\rowcolor{rowshade}
BlurGuard~\cite{kim2025blurguard} & NeurIPS'25 & ImageNet-Edit, MagicBrush, InpaintGuardBench, Helen, WikiArt, VGGFace2 & SD1.4 & \cmark & \cmark & \cmark & \cmark \\
PCA~\cite{guo2025pca} & TIFS'25 & ImageNet, LAION & SD1.4, SD1.5 & \cmark & \xmark & \cmark & \cmark \\
\rowcolor{rowshade}
DeContext~\cite{shen2025decontext} & arXiv'25 & VGGFace2, CelebA-HQ & FLUX.1-Kontext-dev & \xmark & \cmark & \xmark & \xmark \\
SDA~\cite{he2025structure} & arXiv'25 & 100 face/mask pairs + COCO & SD2 & \xmark & \cmark & \cmark & \cmark \\
\rowcolor{rowshade}
Anti-Inpainting~\cite{guo2025antiinpainting} & arXiv'25 & InpaintGuardBench, CelebA-HQ & SD1.5 & \xmark & \xmark & \cmark & \cmark \\
DiffVax~\cite{ozden2026diffvax} & ICLR'26 & CCP dataset & SD Inpainting, IP2P, MagicBrush & \cmark & \cmark & \cmark & \cmark \\
\rowcolor{rowshade}
Universal Immunization~\cite{lee2026universal} & ECCV'26 & Synthetic dataset, ImageNet-Edit & SD1.4, SD1.5, SD2.0, IP2P & \cmark & \cmark & \cmark & \cmark \\
\grouprowImg{bandImg}{Face-swap / biometric}
FaceShield~\cite{jeong2024faceshield} & ICCV'25 & CelebA-HQ, VGGFace2-HQ & DiffFace, DiffSwap, FaceSwap, IP-Adapter & \cmark & \cmark & \cmark & \cmark \\
\rowcolor{rowshade}
FaceSwapGuard \cite{wang2025faceswapguard} & arXiv'25 & CelebA-HQ & FaceShifter, SimSwap & \cmark & \cmark & \cmark & \cmark \\
My Face Is Mine~\cite{yam2025myfaceismine} & arXiv'25 & CelebA-HQ & FaceAdapter & \cmark & \cmark & \cmark & \cmark \\
\rowcolor{rowshade}
AEGIS~\cite{li2026aegis} & arXiv'26 & CelebA, FFHQ, LFW & StarGAN, AttGAN, HiSD & \cmark & \cmark & \cmark & \cmark \\
\grouprowImg{bandDef}{Defenses: Purification}
DiffPure~\cite{diffpure} & ICML'22 & CIFAR-10, CelebA-HQ, ImageNet & Classification models & \cmark & \xmark & N/A & \xmark \\
\rowcolor{rowshade}
GrIDPure~\cite{gridpure} & CVPR'24 & CelebA-HQ, WikiArt & AdvDM, AntiDB, Glaze & \cmark & \cmark & N/A & \xmark \\
PDM-Pure~\cite{xue2024pixel} & arXiv'24 & ImageNet, SDS & AdvDM, SDS, PhotoGuard, Mist & \cmark & \xmark & N/A & \xmark \\
\rowcolor{rowshade}
Noisy Upscaling~\cite{honig2024adversarial} & ICLR'25 & WikiArt, artists & AntiDB, Glaze, Mist & \cmark & \cmark & N/A & \xmark \\
\grouprowImg{bandDef}{Defenses: Unlearning, Red-Teaming}
SLD~\cite{schramowski2023safe} & CVPR'23 & Inappropriate Image Prompts & SD1.4 & N/A & \cmark & \cmark & \xmark \\
\rowcolor{rowshade}
AdvUnlearn~\cite{zhang2024advunlearn} & NeurIPS'24 & I2P, Imagenette, COCO & SD1.4 & N/A & \cmark & \cmark & \cmark \\
GuardT2I~\cite{yang2024guardt2i} & NeurIPS'24 & I2P, I2P-Sexual, SneakyPrompt, MMA-Diffusion, Ring-A-Bell, P4D & SD1.5 & N/A & \cmark & \cmark & \cmark \\
\rowcolor{rowshade}
EraseDiff~\cite{wu2024erasediff} & arXiv'24 & CIFAR10, UTKFace, CelebA, CelebA-HQ & conditional DDIM, unconditional DDIM/DDPM & N/A & \cmark & N/A & \cmark \\
Erasing~\cite{fuchi2024erasing} & arXiv'24 & author-chosen concept with CLIP ImageNet Template & SD1.5 & N/A & \cmark & N/A & \xmark \\
\rowcolor{rowshade}
CAT~\cite{peng2025catadvtrain} & ICML'25 & CelebA-HQ, VGGFace2, WikiArt & AdvDM, Mist, SDS, Glaze, AntiDB, MetaCloak & N/A & \xmark & N/A & \xmark \\
GuardDoor~\cite{zeng2025guarddoor} & arXiv'25 & CelebA, WikiArt & SD2.1 & \cmark & \xmark & N/A & \cmark \\
\rowcolor{rowshade}
DiffShortcut~\cite{liu2024rethinkred} & KDD'26 & VGGFace2 & MetaCloak, AdvDM, PhotoGuard, Glaze & N/A & \xmark & N/A & \xmark \\
\end{longtable}
}

%% file: tables/table4_methods_video.tex
\begin{table}[t]
\centering
\caption{\textbf{Video attack and defense methods.}}
\label{tab:methods_video}
\vspace{-4pt}
{\fontsize{5.0pt}{5.7pt}\selectfont
\setlength{\tabcolsep}{2.2pt}
\renewcommand{\arraystretch}{0.95}
\setlength{\tblrest}{\dimexpr\linewidth-1.50cm-1.36cm-14\tabcolsep\relax}
\begin{tabular}{@{}
>{\rowstrut\raggedright\arraybackslash}m{0.22\tblrest}
>{\centering\arraybackslash}m{1.50cm}
>{\raggedright\arraybackslash}m{0.45\tblrest}
>{\raggedright\arraybackslash}m{0.33\tblrest}
>{\centering\arraybackslash}m{0.34cm}
>{\centering\arraybackslash}m{0.34cm}
>{\centering\arraybackslash}m{0.34cm}
>{\centering\arraybackslash}m{0.34cm}
@{}}
\toprule
\textbf{Method} & \textbf{Venue} & \textbf{Evaluation Sources}
& \textbf{Target Models}
& \textbf{IA} & \textbf{FA} & \textbf{TR} & \textbf{RB} \\
\midrule
\grouprowImg{bandVid}{I2V}
UVCG~\cite{li2024uvcg} & arXiv'24 & DAVIS subset & SVD, TokenFlow, Text2Video-Zero & \cmark & \xmark & \cmark & \xmark \\
\rowcolor{rowshade}
I2VGuard~\cite{gui_i2vguard_nodate} & CVPR'25 & Manually curated & SVD, CogVideoX, ControlNeXt & \cmark & \xmark & \xmark & \xmark \\
DORMANT~\cite{zhou2025dormant} & USENIX'25 & TikTok, Champ, UBC Fashion, TED Talks & Animate Anyone, MagicAnimate, MagicPose, MusePose, Champ, MuseV, UniAnimate, ControlNeXt & \cmark & \cmark & \cmark & \cmark \\
\rowcolor{rowshade}
Vid-Freeze~\cite{chowdhury2026vidfreeze} & arXiv'25 & Manually curated & SVD, CogVideoX & \cmark & \xmark & \xmark & \cmark \\
Anti-I2V~\cite{vu2026antii2v} & CVPR'26 & CelebV-Text, UCF101 & CogVideoX, OpenSora, DynamiCrafter & \cmark & \xmark & \cmark & \cmark \\
\rowcolor{rowshade}
Immune2V~\cite{long2026immune2v} & arXiv'26 & DAVIS subset & Wan, DynamiCrafter, I2VGen-XL & \cmark & \cmark & \cmark & \cmark \\
\grouprowImg{bandVid}{Talking-head}
Silencer~\cite{gan2025silencer} & CVPR'25 & CelebA-HQ, TalkingHead-1KH & Hallo & \cmark & \xmark & \cmark & \cmark \\
\rowcolor{rowshade}
SyncBreaker~\cite{zhang2026syncbreaker} & arXiv'26 & CelebA-HQ--LibriSpeech, HDTF & Hallo & \cmark & \xmark & \xmark & \cmark \\
\grouprowImg{bandVid}{V2V editing}
PRIME~\cite{li2024prime} & arXiv'24 & VIOLENT & TokenFlow, Rerender-A-Video & \cmark & \xmark & \cmark & \cmark \\
\rowcolor{rowshade}
VGMShield~\cite{pang2025vgmshield} & arXiv'24/25 & WebVid-10M, InternVid & Hotshot-XL, I2VGen-XL, LaVie, SEINE, Show-1, SVD, VideoCrafter & \cmark & \cmark & \cmark & \xmark \\
VideoGuard~\cite{cao_videoguard_2025} & arXiv'25 & DAVIS subset & Tune-A-Video, Fate-Zero, Video-P2P & \cmark & \cmark & \xmark & \xmark \\
\grouprowImg{bandVid}{T2V safety}
T2VAttack~\cite{li2025t2vattack} & arXiv'25 & T2VAttackBench & ModelScope, CogVideoX, Open-Sora, HunyuanVideo & \cmark & \cmark & \xmark & \xmark \\
\rowcolor{rowshade}
T2V-OptJail~\cite{liu2025t2voptjail} & arXiv'25 & T2VSafetyBench & Open-Sora, Pika, Luma, Kling & \cmark & \xmark & \cmark & \cmark \\
ConceptGuard~\cite{ma2025conceptguard} & arXiv'25 & ConceptRisk, T2VSafetyBench-TI2V & CogVideoX & N/A & \cmark & \cmark & \cmark \\
\rowcolor{rowshade}
T2VShield~\cite{liang2025t2vshield} & IJCV'26 & T2VSafetyBench, SafeWatch, MSVD & Open-Sora, CogVideoX, Kling, Luma, Pika & N/A & \xmark & \cmark & \cmark \\
\bottomrule
\end{tabular}
}
\end{table}

%% file: tables/table5_methods_3d.tex
\begin{table}[t]
\centering
\setlength{\belowcaptionskip}{0pt}
\caption{\textbf{3D attack and defense methods.}}
\label{tab:methods_3d}
{\fontsize{5.0pt}{5.7pt}\selectfont
\setlength{\tabcolsep}{2.2pt}
\renewcommand{\arraystretch}{0.95}
\setlength{\tblrest}{\dimexpr\linewidth-1.50cm-1.36cm-14\tabcolsep\relax}
\begin{tabular}{@{}
>{\rowstrut\raggedright\arraybackslash}m{0.22\tblrest}
>{\centering\arraybackslash}m{1.50cm}
>{\raggedright\arraybackslash}m{0.45\tblrest}
>{\raggedright\arraybackslash}m{0.33\tblrest}
>{\centering\arraybackslash}m{0.34cm}
>{\centering\arraybackslash}m{0.34cm}
>{\centering\arraybackslash}m{0.34cm}
>{\centering\arraybackslash}m{0.34cm}
@{}}
\toprule
\textbf{Method} & \textbf{Venue} & \textbf{Evaluation Sources}
& \textbf{Target Models}
& \textbf{IA} & \textbf{FA} & \textbf{TR} & \textbf{RB} \\
\midrule
\grouprowImg{band3D}{Editing}
Image2Multiview \cite{sun2024latent} & ICME'25 & Google Scanned Objects, Internet images & Zero123++, Wonder3D & \cmark & \xmark & \cmark & \cmark \\
\rowcolor{rowshade}
DEGauss~\cite{meng2025degauss} & NeurIPS'25 & NeRF-Art, Mip-NeRF 360 & GaussianEditor, DGE, DreamCatalyst, EditSplat & \cmark & \cmark & \xmark & \xmark \\
AdLift~\cite{hong2025adlift} & ICML'26 & IN2N, NeRF-Art, BlendedMVS scenes & IP2P, Instruct-GS2GS, DGE & \cmark & \cmark & \cmark & \cmark \\
\grouprowImg{band3D}{Poisoning / cost}
Poison-splat~\cite{lu2025poison} & ICLR'25 & NeRF-Synthetic, Mip-NeRF 360, Tanks \& Temples & Orig. 3DGS, Scaffold-GS & \cmark & \cmark & \cmark & \cmark \\
\rowcolor{rowshade}
StealthAttack~\cite{ke2025stealthattack} & ICCV'25 & Mip-NeRF 360, Tanks \& Temples, Free & Nerfacto, Instant-NGP & \cmark & \cmark & \xmark & \cmark \\
GaussTrap~\cite{hong2025gausstrap} & arXiv'25 & Blender, Mip-NeRF 360 & Orig. 3DGS & \cmark & \xmark & \xmark & \cmark \\
\grouprowImg{bandDef}{Defenses}
RDSplat~\cite{zhao2025rdsplat} & arXiv'25 & Blender, LLFF, Mip-NeRF 360, IN2N & DiffEdit, InstructPix2Pix & \cmark & \cmark & \xmark & \cmark \\
\rowcolor{rowshade}
RemedyGS~\cite{li2025remedygs} & CVPR'26 & NeRF-Synthetic, Mip-NeRF 360, Tanks \& Temples & Orig. 3DGS, Scaffold-GS & N/A & \xmark & \cmark & \cmark \\
\bottomrule
\end{tabular}
}
\end{table}

%% file: sections/04_evaluation.tex
\section{Evaluation}
\label{sec:evaluation}

Evaluation is multi-objective: an attack must be \emph{imperceptible}, \emph{effective}, \emph{transferable}, \emph{robust} to purification, and \emph{efficient}, and a defense must satisfy all of these while preserving utility. We briefly consolidate the datasets and benchmarks, target models, and metrics used across the surveyed literature.

\paragraph{Datasets and benchmarks.}
Table~\ref{tab:dataset_overview} categorizes every evaluation source appearing in Tables~\ref{tab:methods_image}--\ref{tab:methods_3d} by modality and by its role in attack/defense evaluation. In short, image works pair face datasets (CelebA(-HQ)~\cite{liu2015deep}, VGGFace2~\cite{cao2018vggface2}, FFHQ~\cite{karras2019style}) and art corpora (WikiArt~\cite{tan2018improved}, ArtBench-10~\cite{liao2022artbench}) with general vision corpora (ImageNet~\cite{russakovsky2015imagenet}, COCO~\cite{lin2014microsoft}), dedicated editing and inpainting benchmarks (MagicBrush~\cite{zhang2023magicbrush}, EditBench~\cite{wang2023imagen}, InpaintGuardBench~\cite{choi2024diffusionguard}, Defense-Edit~\cite{zheng2025antidiffusion}), and safety prompt sets for guardrails and unlearning (I2P~\cite{schramowski2023safe}, MMA-Diffusion~\cite{yang2024mma}, Ring-A-Bell~\cite{tsai2024ring}, P4D~\cite{chin2024prompting4debugging}). Video works reuse segmentation and action sources (DAVIS~2017~\cite{pont20172017}, UCF101~\cite{soomro2012ucf101}), human-animation and talking-head corpora (TikTok~\cite{jafarian2021tiktok}, Champ~\cite{zhu2024champ}, CelebV-Text~\cite{yu2023celebv}, HDTF~\cite{zhang2021flow}), and video safety benchmarks (VIOLENT~\cite{li2024prime}, T2VSafetyBench~\cite{miao2024t2vsafetybench}, ConceptRisk~\cite{ma2025conceptguard}). 3D works rely on synthetic objects (NeRF-Synthetic~\cite{mildenhall2021nerf}, Google Scanned Objects~\cite{downs2022google}), real captures (LLFF~\cite{mildenhall2019local}, Mip-NeRF~360~\cite{barron2022mip}, Tanks~\&~Temples~\cite{knapitsch2017tanks}), and instruction-editing testbeds (Instruct-NeRF2NeRF~\cite{haque2023instruct}, NeRF-Art~\cite{wang2023nerf}), with all 3D protocols mandating \emph{novel}-view sampling so that protection is a property of the representation rather than of the optimized views. Many papers additionally evaluate on small author-curated sets (30--2{,}000 items), which aids scenario control but hinders cross-paper comparison, and no standardized benchmark yet exists for 3D content protection---gaps we revisit in Section~\ref{sec:open_challenges}.

\input{tables/table6_dataset_overview}

\paragraph{Target models.}
The Target Models columns of Tables~\ref{tab:methods_image}--\ref{tab:methods_3d} instantiate Eve's downstream capability. Image attacks overwhelmingly target the Stable Diffusion family~\cite{rombach2022high} (v1.x through SDXL), reached through personalization pipelines (DreamBooth~\cite{dreambooth}, Textual Inversion~\cite{gal2022textual}, LoRA variants, Custom Diffusion~\cite{kumari2023multi}), editors and inpainters (InstructPix2Pix~\cite{instructpix2pix}, SDEdit~\cite{meng2022sdedit}, MasaCtrl~\cite{cao2023masactrl}, DiffEdit~\cite{couairondiffedit}, SD~Inpainting), adapters (T2I-Adapter~\cite{mou2024t2i}, IP-Adapter~\cite{ye2023ip}), and, recently, DiT-based editors (FLUX.1~Kontext~\cite{labs2025flux}); face-swap cloaks additionally cover diffusion-based swappers (DiffFace~\cite{kim2025diffface}, DiffSwap~\cite{zhao2023diffswap}, Face-Adapter~\cite{han2024face}) and GAN-based manipulators (FaceShifter~\cite{li2019faceshifter}, SimSwap~\cite{chen2020simswap}, StarGAN~\cite{choi2018stargan}, AttGAN~\cite{he2019attgan}, HiSD~\cite{li2021image}), with hosted services (Replicate, NovelAI) probed as black boxes. Video protections target I2V and T2V generators (SVD~\cite{blattmann2023stable}, CogVideoX~\cite{yang_cogvideox_2024}, DynamiCrafter~\cite{xing2024dynamicrafter}, Open-Sora~\cite{zheng2024open}, I2VGen-XL~\cite{zhang2023i2vgen}, Wan~\cite{wan2025wan}, HunyuanVideo~\cite{wu2025hunyuanvideo}), pose-driven human animators (Animate Anyone~\cite{hu2024animate}, MagicAnimate~\cite{xu2024magicanimate}, MagicPose~\cite{chang2024magicpose}, Champ~\cite{zhu2024champ}, UniAnimate~\cite{wang2025unianimate}), audio-driven talking heads (Hallo~\cite{xu_hallo_2024}, EchoMimic~\cite{chen2025echomimic}), zero-shot video editors (TokenFlow~\cite{geyer2024tokenflow}, Text2Video-Zero~\cite{khachatryan2023text2video}, Tune-A-Video~\cite{wu2023tune}, Fate-Zero~\cite{qi2023fatezero}, Video-P2P~\cite{liu2024video}, Rerender-A-Video~\cite{yang2023rerender}), and commercial systems (Pika, Luma, Kling~\cite{kuaishou2024kling}). 3D methods attack multi-view diffusion and single-image-to-3D reconstruction (Zero123++~\cite{shi2023zero123++}, Wonder3D~\cite{long2024wonder3d}), instruction-guided 3DGS/NeRF editors (GaussianEditor~\cite{wang2024gaussianeditor}, DGE~\cite{chen2024dge}, DreamCatalyst~\cite{kimdreamcatalyst}, EditSplat~\cite{lee2025editsplat}, Instruct-GS2GS~\cite{igs2gs}), and reconstruction backbones (3DGS~\cite{kerbl20233d}, Scaffold-GS~\cite{lu2024scaffold}, Instant-NGP~\cite{muller2022instant}).

\paragraph{Metrics.}
We group the reported metrics into six recurring categories; the middle four correspond to the IA, FA, TR, and RB columns of Tables~\ref{tab:methods_image}--\ref{tab:methods_3d}, while effectiveness and efficiency are reported by virtually every work. Table~\ref{tab:metric_definitions} formalizes the individual metrics---their inputs, outputs, what they measure, and the direction indicating successful protection. \emph{Effectiveness} compares generations from protected inputs against those from clean inputs: FID, LPIPS, SSIM/PSNR, and CLIP similarity for images, complemented by FDFR and ISM in face-centric pipelines; optical-flow similarity, frame consistency, lip-sync scores, and attack-success rate for video; and held-out novel-view fidelity with cross-view consistency for 3D. \emph{Imperceptibility} (IA) instead compares the protected \emph{input} against its clean source (PSNR, LPIPS, MS-SSIM, and human or VLM studies). \emph{Failure analysis} (FA) probes where a protection breaks down (atypical content; unseen prompts, masks, or seeds); it has no standardized metric and remains the least consistently reported dimension. \emph{Transferability} (TR) re-measures effectiveness on backbones, samplers, and tasks the cloak was not optimized against, most notably across the U-Net-to-DiT shift. \emph{Robustness} (RB) re-runs the effectiveness protocol after purification (JPEG compression, cropping, additive noise, \textbf{DiffPure}~\cite{diffpure}, \textbf{PDM-Pure}~\cite{xue2024pixel}), instantiating the minimax of Equation~\eqref{eq:minimax}. Finally, \emph{efficiency} reports protection cost per asset, separating per-sample optimizers from amortized generators that are orders of magnitude faster~\cite{ahn2024nearly,ozden2026diffvax}.

\input{tables/table7_metric_definitions}

%% file: tables/table6_dataset_overview.tex
\providecommand{\dsband}[2]{\multicolumn{3}{@{}l@{}}{\cellcolor{#1}\rowstrut\textbf{#2}}\\}
{\fontsize{6.5pt}{7.8pt}\selectfont
\setlength{\tabcolsep}{2.5pt}
\renewcommand{\arraystretch}{1.12}
\setlength{\LTleft}{\fill}\setlength{\LTright}{\fill}
\setlength{\LTpre}{6pt}\setlength{\LTpost}{6pt}
\setlength{\tblrest}{\dimexpr\linewidth-4\tabcolsep\relax}
\begin{longtable}{@{}
>{\raggedright\arraybackslash}p{0.25\tblrest}
>{\raggedright\arraybackslash}p{0.45\tblrest}
>{\raggedright\arraybackslash}p{0.30\tblrest}
@{}}
\caption{\textbf{Datasets and benchmarks used across the surveyed literature}, grouped by modality and by their role in attack/defense evaluation. Detailed descriptions follow in the text.}\label{tab:dataset_overview}\\[1pt]
\toprule
\textbf{Dataset} & \textbf{Content} & \textbf{Scale} \\
\midrule
\endfirsthead
\multicolumn{3}{@{}l}{\emph{Table~\ref{tab:dataset_overview} continued.}} \\
\toprule
\textbf{Dataset} & \textbf{Content} & \textbf{Scale} \\
\midrule
\endhead
\midrule
\multicolumn{3}{r@{}}{\emph{(continued)}} \\
\endfoot
\bottomrule
\endlastfoot
\dsband{bandImg}{Image}
\dsband{gray!13}{Faces \& identity: source portraits for personalization and face-swap cloaks}
CelebA~\cite{liu2015deep} & in-the-wild celebrity faces; 40 binary attributes, 5 landmarks & 202{,}599 imgs, 10{,}177 IDs \\
VGGFace2~\cite{cao2018vggface2} & celebrity/public-figure faces; large pose and age variation & 3.31M imgs, 9{,}131 IDs \\
FFHQ~\cite{karras2019style} & high-quality aligned Flickr faces; wide demographic coverage & 70{,}000 imgs at $1024^2$ \\
LFW~\cite{huang2008labeled} & unconstrained faces for pair-matching verification & 13{,}233 imgs, 5{,}749 people \\
Helen~\cite{le2012interactive} & high-resolution portraits with dense landmark annotations & 2{,}330 imgs \\
UTKFace~\cite{zhifei2017cvpr} & single faces with age, gender, and ethnicity labels & 23{,}705 imgs \\
\dsband{gray!13}{Art \& style: artworks for style-mimicry and copyright protection}
WikiArt~\cite{tan2018improved} & fine-art paintings; style, genre, and artist labels & 81{,}444 paintings, 27 styles \\
ArtBench-10~\cite{liao2022artbench} & class-balanced, cleanly annotated artworks & 60{,}000 imgs, 10 styles \\
Anita~\cite{Anita2024} & hand-drawn industrial cartoon keyframes (sketch/color/composition) & 16{,}000+ frames, 14 works \\
\dsband{gray!13}{Subject-driven personalization: canonical subjects and evaluation prompts}
DreamBooth~\cite{dreambooth} & casual photos of unique objects and pets, with prompt protocol & 30 subjects, 15 classes \\
\dsband{gray!13}{General vision corpora: broad content for editing and purification evaluation}
ImageNet (ILSVRC)~\cite{russakovsky2015imagenet} & natural object photographs & 1.28M train imgs, 1{,}000 classes \\
LSUN~\cite{yu2015lsun} & large-scale scene and object images & $\sim$10M scene + 59M object imgs \\
LAION-5B~\cite{schuhmann2022laion} & CLIP-filtered web image--text pairs (pre-training corpus) & 5.85B pairs \\
COCO~\cite{lin2014microsoft} & everyday scenes; instance masks and captions & 330K imgs, 80 object categories \\
CIFAR-10~\cite{krizhevsky2009learning} & small natural images, single prominent object & 60{,}000 imgs at $32^2$, 10 classes \\
CCP~\cite{yang2014clothing} & street-fashion photos with clothing parsing & 2{,}098 imgs, 59 tags \\
\dsband{gray!13}{Editing \& inpainting benchmarks: standardized edit tasks and protection testbeds}
InstructPix2Pix~\cite{instructpix2pix} & synthetic instruction-based editing triplets & 454{,}445 triplets \\
MagicBrush~\cite{zhang2023magicbrush} & manually annotated real-image edit sessions & 10{,}388 edit turns \\
EditBench~\cite{wang2023imagen} & text-guided inpainting; masks with three prompt types each & 240 imgs, 720 prompts \\
TEdBench++~\cite{brack2024ledits++} & revised text-editing benchmark with challenging edit types & 120 image--instruction pairs \\
OmniEdit~\cite{wei2025omniedit} & high-resolution synthetic edits across seven task types & 1.2M triplets \\
PPR10K~\cite{liang2021ppr10k} & raw portraits with expert-retouched ground truths & 11{,}161 photos, 1{,}681 groups \\
Defense-Edit~\cite{zheng2025antidiffusion} & protection evaluation against MasaCtrl/DiffEdit editing & 50 image--prompt pairs \\
InpaintGuardBench~\cite{choi2024diffusionguard} & inpainting protection with seen/unseen masks and prompts & 42 imgs, 2{,}100 edit tasks \\
\dsband{gray!13}{Safety \& red-teaming prompt sets: unsafe-generation probes for guardrails and unlearning}
I2P~\cite{schramowski2023safe} & real-world inappropriate prompts from lexica.art & 4{,}703 prompts, 7 categories \\
SneakyPrompt~\cite{yang2024sneakyprompt} & NSFW jailbreak prompts plus benign substitutes & 200 + 100 prompts \\
MMA-Diffusion~\cite{yang2024mma} & multimodal red-teaming: adversarial prompts and images & 1{,}030 prompts + 60 adv.\ imgs \\
Ring-A-Bell~\cite{tsai2024ring} & concept-recall prompts against erasure (nudity/violence) & 95 + 250 prompts \\
P4D~\cite{chin2024prompting4debugging} & optimized jailbreak prompts built on I2P & 4{,}703 prompts + 2 object classes \\
\dsband{bandVid}{Video}
\dsband{gray!13}{General video \& action: segmentation and recognition sources reused for protection}
DAVIS 2017~\cite{pont20172017} & densely annotated video object segmentation & 150 seqs, 10{,}459 frames \\
UCF101~\cite{soomro2012ucf101} & YouTube action-recognition clips & 13{,}320 clips, 101 actions \\
\dsband{gray!13}{Human animation \& motion: pose-driven animation protection}
TikTok~\cite{jafarian2021tiktok} & single-person dance sequences with masks and UV coords & 300+ seqs, $>$100K frames \\
Champ curated dataset~\cite{zhu2024champ} & in-the-wild human videos for image animation & $\sim$5{,}000 videos, 1M frames \\
UBC Fashion~\cite{zablotskaia2019dwnet} & fashion-model videos, static camera & 600 videos, $\sim$210K frames \\
TED-talks~\cite{siarohin2021motion} & cropped upper-body speakers on stage & 411 videos, 1{,}322 chunks \\
\dsband{gray!13}{Talking-head \& face video: audio-driven and face-centric generation}
CelebV-Text~\cite{yu2023celebv} & face clips paired with rich text descriptions & 70{,}000 clips, 1.4M texts \\
TalkingHead-1KH~\cite{wang2021one} & high-quality talking-head footage & $\sim$180K videos, $\sim$1{,}000 h \\
HDTF~\cite{zhang2021flow} & high-definition talking faces & 362 videos, 15.8 h, 300+ subjects \\
\dsband{gray!13}{T2V corpora: text--video training and evaluation sources}
OpenVid-1M~\cite{nan2025openvid} & curated high-quality text--video pairs & 1.02M clips, 2{,}051 h \\
InternVid~\cite{wang2024internvid} & web-scale video--text with generated captions & 234M clips from 7M videos \\
MSVD~\cite{chen-dolan-2011-collecting} & short clips with many parallel captions & 2{,}089 clips, 85{,}550 sentences \\
\dsband{gray!13}{Safety \& attack benchmarks: malicious editing and unsafe-generation probes}
VIOLENT~\cite{li2024prime} & malicious-editing testbed of well-known identities & 35 clips, 10 identities \\
T2VAttackBench~\cite{li2025t2vattack} & semantic/temporal adversarial prompts (VBench-derived) & 157 prompts (105 semantic, 52 temporal) \\
T2VSafetyBench~\cite{miao2024t2vsafetybench} & malicious T2V prompts across safety dimensions & 4{,}400 prompts, 12 categories \\
SafeWatch-Bench~\cite{chen2025safewatch} & video-guardrail corpus, real and generated & 2M+ clips, 6 categories \\
ConceptRisk~\cite{ma2025conceptguard} & TI2V multimodal risk instances with safe rewrites & 8{,}000 instances, 200 concepts \\
\dsband{band3D}{3D}
\dsband{gray!13}{Synthetic objects: controlled, noise-free reconstruction sources}
NeRF-Synthetic~\cite{mildenhall2021nerf} & path-traced Blender scenes, object-centric $360^\circ$ views & 8 scenes, 400 views each \\
Google Scanned Objects~\cite{downs2022google} & photorealistic 3D-scanned household items & 1{,}000+ assets \\
\dsband{gray!13}{Real captures: novel-view synthesis under realistic conditions}
LLFF~\cite{mildenhall2019local} & forward-facing handheld captures, sparse views & 8 scenes, 20--62 imgs each \\
Mip-NeRF 360~\cite{barron2022mip} & unbounded $360^\circ$ indoor/outdoor scenes & 9 scenes, $\sim$100--330 imgs \\
Tanks \& Temples~\cite{knapitsch2017tanks} & large real scenes with laser-scanned ground truth & video-based captures, 7 training + 14 benchmark test scenes \\
F$^2$-NeRF Free~\cite{wang2023f2} & unbounded scenes with free camera trajectories & scene-organized captures, 7 scenes \\
BlendedMVS~\cite{yao2020blendedmvs} & blended-render multi-view-stereo scenes & 113 scenes, 17K+ imgs \\
\dsband{gray!13}{Editing testbeds: instruction- and text-driven 3D editing}
IN2N scenes~\cite{haque2023instruct} & real Nerfstudio scenes for instruction-driven editing & $360^\circ$ scenes, faces, portraits \\
NeRF-Art scenes~\cite{wang2023nerf} & captured multi-view scenes for text-driven stylization & self-portrait captures + H3DS + LLFF \\
\end{longtable}
}

%% file: tables/table7_metric_definitions.tex
{\fontsize{6.5pt}{7.8pt}\selectfont
\setlength{\tabcolsep}{2.5pt}
\renewcommand{\arraystretch}{1.22}
\setlength{\LTleft}{\fill}\setlength{\LTright}{\fill}
\setlength{\LTpre}{6pt}\setlength{\LTpost}{6pt}
\setlength{\tblrest}{\dimexpr\linewidth-6\tabcolsep\relax}
\begin{longtable}{@{}
>{\raggedright\arraybackslash}p{0.18\tblrest}
>{\raggedright\arraybackslash}p{0.26\tblrest}
>{\raggedright\arraybackslash}p{0.48\tblrest}
>{\centering\arraybackslash}p{0.08\tblrest}
@{}}
\caption{\textbf{Definitions of the evaluation metrics.} Each metric is written as a function of its inputs. $x$ is Alice's clean asset and $x\oplus\delta$ its protected version; $c$ is the conditioning (prompt, subject, or audio $a$); $y=D_\theta(z,\,x\oplus\delta,\,c)$ is Eve's output on the protected asset and $y^{\mathrm{clean}}=D_\theta(z,\,x,\,c)$ her output on the clean asset; $V$ is a generated video and $\{\mathcal{R}_v\}_v$ are renders of a 3D asset from viewpoints $v$. The \textbf{Goal} column gives the direction indicating \emph{successful protection}: degraded generation (top block), an invisible cloak (middle block), and cheap protection (bottom block).}\label{tab:metric_definitions}\\[1pt]
\toprule
\textbf{Metric} & \textbf{Signature (inputs $\to$ output)} & \textbf{What it measures} & \textbf{Goal} \\
\midrule
\endfirsthead
\multicolumn{4}{@{}l}{\emph{Table~\ref{tab:metric_definitions} continued.}} \\
\toprule
\textbf{Metric} & \textbf{Signature (inputs $\to$ output)} & \textbf{What it measures} & \textbf{Goal} \\
\midrule
\endhead
\midrule
\multicolumn{4}{r@{}}{\emph{(continued)}} \\
\endfoot
\bottomrule
\endlastfoot
\grouprow{gray!20}{Output side: effectiveness (Eve's generation should degrade)} \\
FID & $(\{y\},\,\{y^{\mathrm{ref}}\})\to\mathbb{R}_{\ge 0}$ & Fr\'echet distance between Gaussians fitted to Inception features of the generated and reference sets; distribution-level realism & $\uparrow$ \\
LPIPS & $(y,\,y^{\mathrm{clean}})\to\mathbb{R}_{\ge 0}$ & weighted distance between deep-network activations of an image pair; perceptual dissimilarity & $\uparrow$ \\
SSIM\,/\,PSNR & $(y,\,y^{\mathrm{clean}})\to[0,1]$\,/\,dB & structural similarity (luminance, contrast, structure) / log-scaled peak-signal-to-MSE ratio; low-level fidelity & $\downarrow$ \\
CLIP-I\,/\,CLIP-T & $(y,\,x)$ or $(y,\,c)\to[-1,1]$ & cosine similarity of CLIP embeddings; semantic faithfulness of the output to the source subject (-I) or to the prompt (-T) & $\downarrow$ \\
FDFR & $\{y\}\to[0,1]$ & fraction of generated images in which a face detector finds no face & $\uparrow$ \\
ISM & $(y,\,x)\to[-1,1]$ & cosine similarity of face-recognition embeddings between the generated face and the protected identity & $\downarrow$ \\
Caption-Sim. & $(\mathrm{cap}(y),\,c)\to[0,1]$ & text similarity between an automatic caption of the output and the editing prompt; whether the intended edit semantically succeeded & $\downarrow$ \\
Semantic-IoU & $(m_y,\,m_{\mathrm{ref}})\to[0,1]$ & overlap between the mask of the edited concept in the output ($m_y$) and its expected support ($m_{\mathrm{ref}}$) & $\downarrow$ \\
Flow similarity & $(V,\,V^{\mathrm{clean}})\to\mathbb{R}$ & agreement of optical-flow fields; preservation of the intended motion & $\downarrow$ \\
Frame consist. & $V\to[-1,1]$ & mean feature (e.g., CLIP) similarity of adjacent frames; temporal coherence of the video & $\downarrow$ \\
Sync-C\,/\,Sync-D & $(V,\,a)\to\mathbb{R}$ & SyncNet confidence / embedding distance between lip motion and the driving audio & $\downarrow$\,/\,$\uparrow$ \\
ASR & $\{(c_i,\,y_i)\}_i\to[0,1]$ & fraction of attempts eliciting the adversary's intended (e.g., unsafe) generation; reported from the guardrail's perspective & $\downarrow$ \\
View consist. & $\{\mathcal{R}_v\}_v\to\mathbb{R}$ & cross-view agreement of the edited 3D content (e.g., pairwise perceptual distance across viewpoints) & $\downarrow$ \\
\midrule
\grouprow{gray!20}{Input side: imperceptibility (the cloak should stay invisible)} \\
PSNR\,/\,MS-SSIM & $(x\oplus\delta,\,x)\to$ dB\,/\,$[0,1]$ & pixel fidelity / multi-scale structural similarity of the protected asset to its clean source & $\uparrow$ \\
LPIPS & $(x\oplus\delta,\,x)\to\mathbb{R}_{\ge 0}$ & perceptual visibility of the cloak itself & $\downarrow$ \\
Human\,/\,VLM study & $(x\oplus\delta,\,x)\to$ score & rater (or GPT-4o) detectability or preference between protected and clean assets & chance \\
\midrule
\grouprow{gray!20}{Cost: efficiency (protection should be affordable)} \\
GPU-hours\,/\,runtime & $(\text{pipeline},\,x)\to$ time & accelerator or wall-clock cost to protect one asset; amortized methods pay one-off training plus per-asset inference & $\downarrow$ \\
\end{longtable}
}

%% file: sections/05_open_challenges_conclusion.tex
\section{Open Challenges and Conclusion}
\label{sec:open_challenges}

Despite rapid progress, the adversarial robustness of diffusion models remains an open problem. We distill the primary challenges cutting across image, video, and 3D into three high-leverage directions for future work.

\paragraph{Adaptive countermeasures and evaluation.} Most protective cloaks are vulnerable to a continuous cycle of neutralization by model-side preprocessing or input purification. Because the party deploying the model controls the input pipeline, fixed perturbations are routinely mitigated by adaptive transformations, such as lossy compression, adaptive blurring, or diffusion-based regeneration. Breaking this cycle requires moving away from empirical hardening against known countermeasures toward certified defenses that hold against broad classes of unseen purifiers. Furthermore, progress is obscured by a lack of standardized benchmarks. Evaluating against static baselines frequently overstates real-world protection; robust evaluation must integrate common datasets, threat models, and adaptive purifier suites to ensure apparent progress can be trusted.

\paragraph{Cross-model and cross-sample generalization.} Current immunizers typically rely on test-time optimization tailored to a specific white-box backbone and a single asset. This process is computationally expensive and transfers poorly to real-world commercial services, where model weights, architectures, and preprocessing steps are completely hidden behind black-box APIs. To be practical, cloaks must attack components shared across multiple pipelines. Simultaneously, future work must resolve the tension between per-sample strength and cross-sample speed. Developing universal perturbations or amortized generators capable of protecting arbitrary unseen content in a single forward pass---without sacrificing the robustness of bespoke optimization---remains a critical open problem.

\paragraph{Theoretical foundations and modality expansion.} While research is heavily concentrated on the image domain, extending protection to video and 3D assets---arguably the higher-stakes settings for real-world misuse, such as deepfakes---remains an urgent frontier. These advanced modalities introduce complex structural constraints: video adds a temporal axis that dilutes per-frame cloaks alongside aggressive compression codecs, while 3D requires perturbations that survive multi-view rendering pipelines without being smoothed away. Overcoming these challenges requires transitioning from empirical trial-and-error to a principled, theoretical characterization of the adversarial perturbation space. A systematic mapping of which geometric and semantic directions fundamentally survive purification and multi-modal transformations would turn the current empirical patchwork into a predictive science.

\paragraph{Conclusion.} We surveyed adversarial attacks and defenses for diffusion models across image, video, and 3D under a unified, task-centric framework. The literature demonstrates that static, single-target protections are no longer sufficient in an environment of adaptive countermeasures. Sustained progress in content immunization now depends on developing defenses that generalize across diverse models and samples, establishing standardized adaptive evaluation metrics, and scaling robust protective frameworks to under-studied video and 3D domains.